%% file: acl_latex.tex
\pdfoutput=1

\documentclass[11pt]{article}

\usepackage[final]{acl}

\usepackage{times}
\usepackage{latexsym}
\usepackage{booktabs}
\usepackage{makecell}

\usepackage[T1]{fontenc}

\usepackage[utf8]{inputenc}
\usepackage{svg}

\usepackage{microtype}

\usepackage{inconsolata}
\usepackage{multirow}
\usepackage[normalem]{ulem}
\useunder{\uline}{\ul}{}

\usepackage{graphicx}

\defcitealias{cs-anli}{CS-ANLI}

\title{Soft Language Prompts for Language Transfer}

\author{Ivan Vykopal$^{1, 2}$, Simon Ostermann$^{3}$ \and Marián Šimko$^{2}$ \\
    $^{1}$ Faculty of Information Technology, Brno University of Technology, Brno, Czech Republic\\
    $^{2}$ Kempelen Institute of Intelligent Technologies, Bratislava, Slovakia \\
    \texttt{\{name.surname\}@kinit.sk} \\
    $^{3}$ German Research Center for Artificial Intelligence (DFKI), Saarbrücken, Germany  \\
  \texttt{simon.ostermann@dfki.de}
  }

\begin{document}
\maketitle
\begin{abstract}

Cross-lingual knowledge transfer, especially between high- and low-resource languages, remains challenging in natural language processing (NLP). This study offers insights for improving cross-lingual NLP applications through the combination of parameter-efficient fine-tuning methods. We systematically explore strategies for enhancing cross-lingual transfer through the incorporation of language-specific and task-specific adapters and soft prompts. We present a detailed investigation of various combinations of these methods, exploring their efficiency across 16 languages, focusing on 10 mid- and low-resource languages. We further present to our knowledge the first use of soft prompts for language transfer, a technique we call \textbf{soft language prompts}. Our findings demonstrate that in contrast to claims of previous work, a combination of language and task adapters does not always work best; instead, combining a soft language prompt with a task adapter outperforms most configurations in many cases.

\end{abstract}

\input{intro}

\input{related_work}

\input{methodology}

\input{experiments}

\input{conclusion}

\bibliography{custom}

\appendix

\input{appendix}

\end{document}

%% file: intro.tex
\section{Introduction}

Many multilingual large language models (LLMs) have been developed in recent years, demonstrating promising performance on various NLP tasks across multiple languages~\cite{xue2021mt5, workshop2023bloom}. These models are pre-trained on extensive corpora of unlabelled data in numerous languages, allowing an adaptation to linguistic characteristics and nuances. In addition, LLMs are often further trained on downstream tasks in a selected subset of languages~\cite{muennighoff2023crosslingual}. However, only few LLMs focus on low-resource languages~\cite{tang2020multilingualtranslationextensiblemultilingual, xue2021mt5, ustun2024aya}. 

As the number of covered languages in the model increases, the issue of the \textit{curse of multilinguality} arises. This problem occurs when the LLM's capacity is limited, causing languages with less training data to perform poorly~\cite{conneau-etal-2020-unsupervised}. Various approaches have been employed to address this limitation, primarily involving additional trainable parameters specific to individual languages~\cite{pfeiffer-etal-2020-mad, pfeiffer-etal-2023-mmt5}.


\begin{figure}
\centering
\hspace*{-0.15cm}\includegraphics[width=1.05\columnwidth]{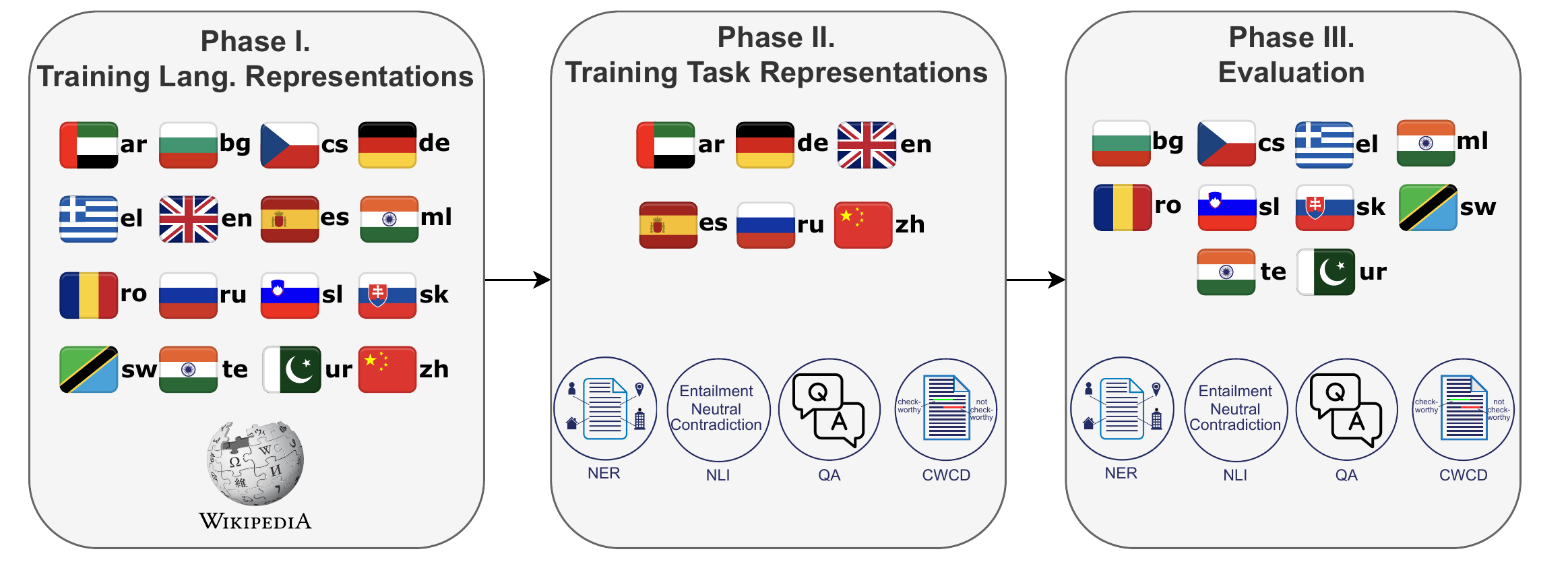}
\caption{The full pipeline consists of training language and task representations along with evaluation on four selected tasks.}
\label{fig:pipeline}
\end{figure}

An alternative to language-specific tuning is \textit{cross-lingual transfer}, where researchers investigate the knowledge transfer between high and low-resource languages. In cross-lingual transfer methods, an LLM is trained on a downstream task in one language, most often high-resource, and evaluated in other languages~\cite{PIKULIAK2021113765}. However, training only task-specific representations does not always capture the nuances of languages on which the LLM has not been trained or has been trained only to a small extent. Therefore, incorporating language-specific features can enhance knowledge transfer across languages. 

Previous work has primarily investigated language and task representations by training language and task-specific adapters~\cite{pfeiffer-etal-2020-mad, parovic-etal-2022-bad} or by employing language arithmetics~\cite{klimaszewski2024train}. Nonetheless, other approaches that involve adding additional parameters to the model for language representation have not been thoroughly explored. This opens the opportunity to explore a combination of language and task representations using other methods and their impact in cross-lingual settings.

To explore the utilization of language and task representations, we evaluate various configurations by combining two parameter-efficient fine-tuning (PEFT) methods that incorporate additional parameters into the LLM, namely \textit{adapters} and \textit{prompt-tuning}. Adding these additional language- and task-specific parameters increases the capacity of an mT0-{\scshape Base} model and improves cross-lingual performance. We evaluate the performance of each configuration by training on six high-resource languages and evaluating its effectiveness on 10 mid- and low-resource languages on four selected tasks\footnote{Code is available at: \url{https://github.com/kinit-sk/adapter-prompt-evaluation}}.
Our main contributions are: 
\begin{itemize}
    \item We propose \textbf{soft language prompts} as an alternative method for cross-lingual transfer.
    \item We comprehensively evaluate combinations of adapters and soft prompts in cross-lingual transfer and find that language prompts provide a viable alternative to language adapters, especially for low-resource languages.
    \item In addition, we provide an exhaustive evaluation of both prompts and adapters for task transfer. We find that the best combination of adapters and prompts for task and language transfer depends highly on task and language, resp., and that no solution clearly outperforms the others.
\end{itemize}

%% file: related_work.tex
\section{Related Work}

\paragraph{Adapters and Soft Prompts.}

PEFT methods are designed to address the problem of the increasing number of trainable parameters in LLMs~\cite{he2022unifiedviewparameterefficienttransfer, NEURIPS2023_1feb8787, zhang2023adalora, xu2023parameterefficient, xie2024discovering}. These methods reduce the number of trained parameters and incorporate new parameters commonly used to train LLMs on other tasks. Adapters~\cite{houlsby2019parameter} and Prompt-Tuning~\cite{lester-etal-2021-power} represent two PEFT methods for adapting LLMs to different NLP domains. Adapters incorporate new parameters into the transformer architecture by including down- and up-projection layers along with residual connection, while prompt-tuning introduced trainable soft-prompts prepended to input embeddings to condition the LLM's generation. 

\paragraph{Limitations of Multilingual LLMs.}

One major limitation of LLMs is \textit{catastrophic forgetting}, which occurs when training the LLM on a new task, causing it to partially or entirely forget previously learned knowledge for other tasks~\cite{MCCLOSKEY1989109, luo2024empirical, ren2024analyzing}. This forgetting extends beyond task-specific knowledge to language-specific knowledge if the model is fine-tuned on a subset of the original languages~\cite{vu-etal-2022-overcoming, 10.1145/3539618.3592043}.

Another challenge with multilingual LLMs is associated with the number of languages on which these LLMs have been pre-trained~\cite{conneau-etal-2020-unsupervised, pfeiffer-etal-2022-lifting}. Previous research has shown that as the number of languages covered by LLMs increases, their performance on various NLP tasks degrades~\cite{pmlr-v119-hu20b, ponti2020xcopa}. Additionally, low-resource languages are often underrepresented during pre-training, resulting in poor performance in these languages~\cite{wu2020languages}.

\paragraph{Cross-Lingual Transfer.}

Given the many low-resource and underrepresented languages, cross-lingual transfer is crucial for training LLMs to address NLP tasks in various languages~\cite{PIKULIAK2021113765}. A common approach involves training LLMs in one language and evaluating them in another. Recent methods use additional parameters to create language-specific representations, assisting LLMs in solving NLP tasks in low-resource languages~\cite{ustun-etal-2020-udapter, ansell-etal-2022-composable}. These include training task adapters on top of language adapters~\cite{pfeiffer-etal-2020-mad, ansell-etal-2021-mad-g, pfeiffer-etal-2023-mmt5, kunz2024impact}, training language adapters on source and target languages~\cite{parovic-etal-2022-bad}, and fusing multiple task~\cite{lee-etal-2022-fad} or language adapters~\cite{rathore-etal-2023-zgul}. Other approaches leverage soft prompts~\cite{huang-etal-2022-zero, philippy-etal-2024-soft} or grammar prompting~\cite{10.5555/3666122.3668959}. While many works focus on specific tasks, our study explored different combinations of adapters and soft prompts for cross-lingual transfer on four tasks, minimizing the reliance on machine translation, which is often unreliable for low-resource languages.

%% file: methodology.tex
\section{Methodology}

We propose a comprehensive study on combinations of language and task representations using adapters and soft prompts. We evaluate for the first time the capabilities of \textbf{soft language prompts} in a systematic manner and evaluate the performance of diverse combinations of prompts and adapters in cross-lingual settings. Our pipeline, consisting of training, evaluation,  multiple languages and tasks that constitute each step, is illustrated in Figure~\ref{fig:pipeline}.

In the following sections, we first give details on methods that we investigate for representing language (Section \ref{sec:Lang-rep}) and task information (Section \ref{sec:Task-rep}). We then explain the combinations of soft prompts and adapters that we evaluate (Section \ref{subsec:combining}).

\subsection{Language Representation}
\label{sec:Lang-rep}

\paragraph{Language Adapters.}

Previous work has investigated the effectiveness of training language-specific transformation using the adapter architecture~\cite{pmlr-v97-houlsby19a}. \citet{pfeiffer-etal-2020-mad} proposed a MAD-X framework, which includes training language adapters using the masked language modeling objective on unlabelled data. Inspired by language adapters proposed by the authors, we build upon their architecture and the approach used to train language adapters. Language adapters in our settings are incorporated into each transformer layer of the LLM and trained using unlabelled data.

\paragraph{Soft Language Prompts.}

Soft Prompt Tuning offers a promising, parameter-efficient method for adapting LLMs. While previous work has predominantly explored task-specific soft prompts aimed at enhancing task transferability, typically focusing on a single language~\cite{vu-etal-2022-spot, asai-etal-2022-attempt}, we extend this approach by training language-specific soft prompts to guide multilingual LLMs toward a target language. Given that multilingual LLMs can generate responses in various languages, we defined a soft language prompt as a set of token embeddings prepended to the input embedding. These embeddings are then fed into the LLM to condition its output to the desired language.
 
Existing studies have highlighted the importance of soft prompt initialization in optimizing the performance of LLMs. \citet{lester-etal-2021-power} outline three possible strategies: (1) \textit{random initialization using a Gaussian distribution}; (2) \textit{initialization from the model's vocabulary}; and (3) \textit{initialization with the embeddings of output classes for classification tasks}. While each method has its strengths and limitations, none are directly applicable to our experiments, which focus on multilingual LLMs. To address this, we introduce a language-specific text instruction for soft prompt initialization (see Appendix~\ref{sec:initialization}). In this approach, the text instruction is first embedded, and if its length is shorter than the required soft prompt size, the embedding is repeated until the desired length is achieved.

\paragraph{Language Modeling Objective.}

Training language-specific representations requires unlabelled data from the selected languages and careful selection of an appropriate training objective. Given our use of an encoder-decoder architecture, we adopt \textit{span corruption} as the training objective, which has been shown to be effective in prior work~\cite{10.5555/3455716.3455856, xue2021mt5}.
Unlike the casual language modeling objective, where the LLM predicts the next token in a sequence, \textit{span corruption} randomly masks 15\% of the tokens in the input text using sentinel tokens. These tokens serve solely to mark the masked parts, which the LLM is tasked to reconstruct~\cite{10.5555/3455716.3455856}. Finally, the LLM is trained to predict the original tokens for the masked portions, enabling it to learn linguistic nuances and patterns that are crucial for training task-specific adapters and soft prompts.

\subsection{Task Representation}
\label{sec:Task-rep}

\paragraph{Task Adapters.}

Similarly to language adapters, we use task-specific adapters, represented by the same architecture, which are incorporated into each transformer layer of the LLM. However, when combining task representations with language representations, the final architecture differs across configurations and depends on the type of language representation used during the training and inference. Detailed information on the architecture for all combinations is in Section~\ref{subsec:combining}. 

Task adapters are updated only during training on the desired downstream task, while the rest of the model, along with the language representation, is kept frozen. In the case of task-specific representations, LLMs learn knowledge that is characteristic of the specified tasks and that should be language-independent.

\paragraph{Soft Task Prompts.}

In addition to task adapters, we also use soft task prompts that employ the same architecture and parameters used for soft language prompts. The difference when using a soft task prompt occurs in the configuration consisting of a soft language prompt and a soft task prompt. With this configuration, both soft prompts are combined using a concatenation operation and further fed into the model to condition the final generation.

\subsection{Evaluated Combinations of Adapters and Soft Prompts}
\label{subsec:combining}

\begin{figure*}[t]
\centering
\includegraphics[width=.9\textwidth]{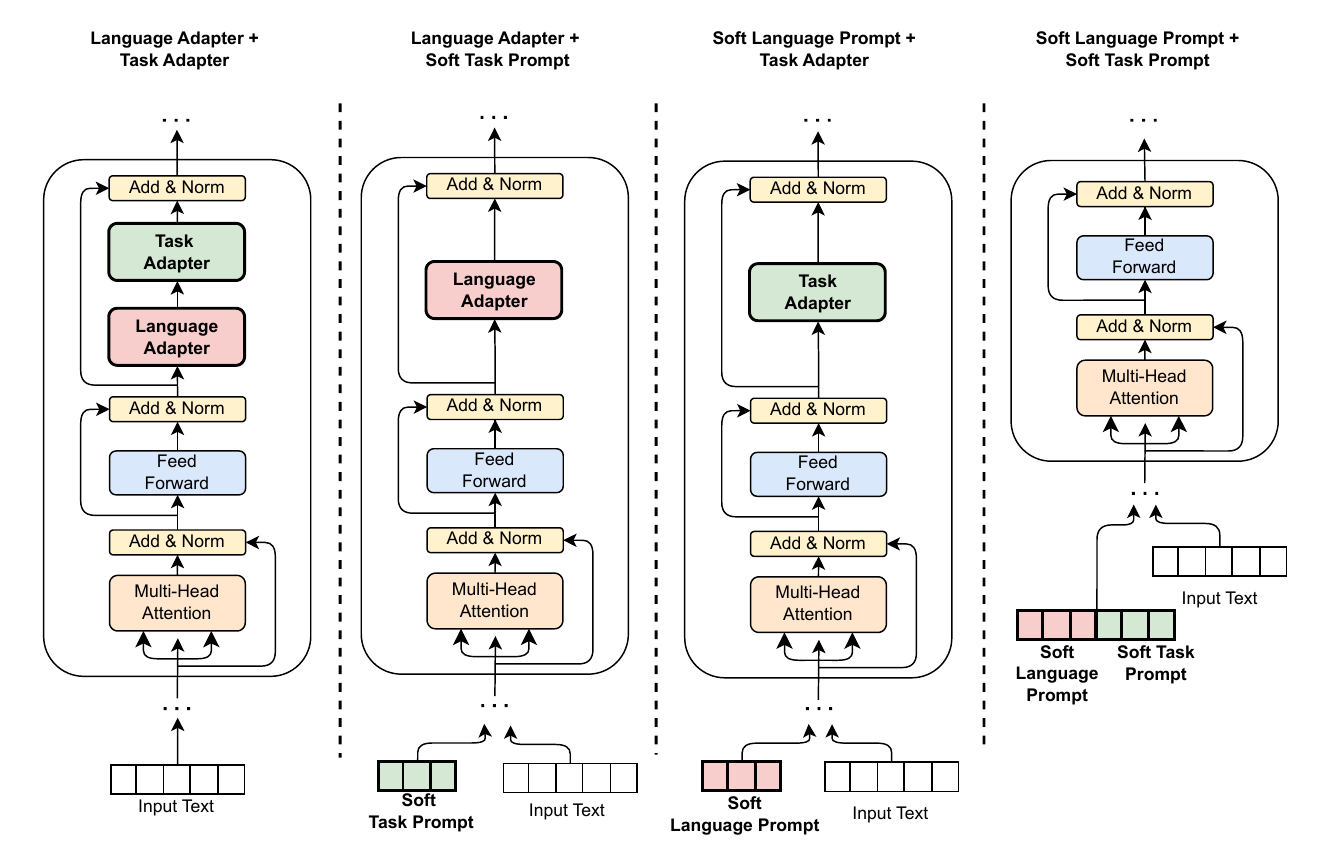}
\caption{The architecture for all combinations of language and task representations in our experiments. These configurations include: (1) Language and Task Adapters; (2) Language Adapter and Soft Task Prompt; (3) Soft Language Prompt and Task Adapter; and (4) Soft Language and Soft Task Prompts. Language representations are in red, while task representations are in green color.}
\label{fig:configurations}
\end{figure*}

Since our experiments are focused on evaluating language and task representations and their combination, we define six possible configurations: (1) only \textit{task adapter}; (2) only \textit{soft task prompt}; (3) MAD-X~\cite{pfeiffer-etal-2020-mad}, i.e. the combination of \textit{language and task adapter}; (4) the combination of \textit{language adapter and soft task prompt}; (5) the combination of \textit{soft language prompt and task adapter}; and (6) the combination of \textit{soft language prompt and soft task prompt}. The position of task representations within the LLM highly depends on the type of language representation used in experiments. The architecture along with the form of the input for all configurations are illustrated in Figure~\ref{fig:configurations}.

\paragraph{Single Task Adapters \& Soft Task Prompts.}

The configurations that employ only task adapters or task soft prompts aim at training task representations, without incorporating language-specific representation. Adapters and soft prompts were trained independently on each selected dataset, and the resulting task representations were evaluated across all defined languages. During this process, only the adapters and soft prompts are trained, while the rest of the LLM remained frozen.

\paragraph{Language Adapters \& Task Adapters.}

Beyond training task representations alone, we also trained a task adapter on top of a pre-trained language adapter, reproducing the approach outlined in MAD-X~\cite{pfeiffer-etal-2020-mad}. Our method utilizes the same architecture but with distinct training hyperparameters, fitted to the tasks at hand. In this setup, the task adapter takes the output of the language adapter as input and further processes it. During training, only the task adapter is trained, while both the language adapter and LLM remain frozen.

\paragraph{Adapters and Soft Prompts Combinations.}

In our study, we introduce two combinations of language and task representation using adapters and soft prompts. The first configuration involves soft task prompts along with a language adapter. This combination incorporates trained language-specific knowledge using a language adapter, and a soft task prompt trained on the desired downstream task.

The second combination includes training a task adapter with the trained soft language prompt. Soft language prompts condition LLMs to activate knowledge specific to the desired language, while task adapters learn task-specific knowledge.

\paragraph{Soft Language Prompts \& Soft Task Prompts.}

The last configuration includes soft language and soft task prompts. Inspired by stacking language and task adapters on top of each other, we concatenated embeddings of language and task prompts to a final soft prompt, with the LLM and soft language prompt being frozen during training.

%% file: experiments.tex
\section{Experiments}

\subsection{Model Selection}

We selected an encoder-decoder architecture, the mT0-{\scshape Base} model, to conduct a cross-lingual evaluation. mT0 is based on the pre-trained multilingual mT5 model, which has been further fine-tuned on a collection of 46 languages across 16 NLP tasks~\cite{muennighoff2023crosslingual}. The model selection played a crucial role in further experiments and we conducted several preliminary experiments with the original mT5-{\scshape Base} model. However, we observed that in the case of using the pre-trained model, which has not been further fine-tuned on downstream tasks, prompt-tuning is not sufficient to guide the LLM to produce meaningful outputs.

\subsection{Languages}

The original mT5 model was pre-trained on over 100 languages, while mT0 employed only 46 for further fine-tuning. From the list of languages supported by mT5, we selected 16 languages and categorized them into two groups: high-  and mid- along with low-resource languages. On the one hand, we consider Arabic, German, English, Spanish, Russian and Chinese to be high-resource languages. On the other hand, we consider Czech, Greek, Romanian and Slovenian as mid-resource and Bulgarian, Malayalam, Slovak, Swahili, Telugu and Urdu as low-resource languages. Our distinction between these two groups is based on the number of resources available for each language (in terms of unlabelled and labelled data).

We included languages from various families (e.g., Indo-European, Dravidian, Sino-Tibetan) and script types in the low-resource category, such as Latin, Arabic, Cyrillic and other non-Latin. The purpose of including multiple scripts and language families in our cross-lingual evaluation is to investigate the ability of the mT0 model to transfer knowledge between more similar and more distant languages, with respect to both script and language features.

To train language representations on unlabelled data, we selected Wikipedia as a source that contains many articles in various languages, including low-resource ones. All Wikipedia data were taken from the latest preprocessed dump from HuggingFace\footnote{\url{https://huggingface.co/datasets/wikimedia/wikipedia}}, November 2023.

\subsection{Tasks}

In order to evaluate the capabilities of mT0-{\scshape Base} for cross-lingual transfer, we choose four distinct tasks involving various NLP areas to explore the model performance. These tasks differ in the type of the provided output and include question answering (QA), named-entity recognition (NER), natural language inference (NLI), and check-worthy claim detection (CWCD). They were selected based on the availability of datasets for selected languages and to include various NLP tasks related to reading comprehension, recognizing textual entailment, or fact-checking. Table~\ref{tab:datasets} lists the datasets used in our experiments. For Bulgarian, there is no question answering dataset available.

Due to the absence of datasets for some languages, we employed Google Translate to translate data for several languages. This concerns, in particular, the dataset for the Slovak NLI task and the dataset for check-worthy claim detection. In the case of the missing Slovak NLI dataset, we utilized the CS ANLI dataset and translated it from Czech to Slovak. For check-worthy claim detection, we translated the English dataset into multiple languages to obtain results for comparison.\footnote{To evaluate the accuracy of the translations, we manually verified a subset of samples, with a particular focus on translations between Czech and Slovak, leveraging input from native speakers. Our analysis found that the translations generated by Google Translate were correct for this language pair.}

\begin{table}[t]
    \centering
    \resizebox{\columnwidth}{!}{%
    \begin{tabular}{lllll}
        \toprule
        \textbf{Dataset} &  \textbf{Task} & \textbf{Languages} & \textbf{Citation} \\
        \midrule
        {\scshape SQuAD} & QA & en & \citet{rajpurkar-etal-2016-squad} \\
        {\scshape MLQA} & QA & ar, de, hi, zh, es, vi  & \citet{lewis2019mlqa} \\
        XQuAD & QA & el, ro & \citet{artetxe-etal-2020-cross} \\
        {\scshape SK-QuAD} & QA & sk & \citet{hladek2023slovak} \\
        {\scshape Czech SQuAD} & QA & cs & \citet{10.1007/978-3-030-58323-1_18} \\
        TeQuAD & QA & te & \citet{vemula-etal-2022-tequad} \\
        KenSWQuAD & QA & sw & \citet{10.1145/3578553} \\
        UQA & QA & ur & \citet{arif-etal-2024-uqa} \\
        Slovene SQuAD & QA & sl & \citet{borovivc2022slovene} \\
        IndicQA & QA & ml & \citet{doddapaneni-etal-2023-towards} \\
        \midrule
        WikiANN & NER & \makecell[tl]{ar, bg, cs, de, el,\\ en, es, ml, ro, ru,\\sl, sk, sw, te, ur, zh} & \citet{rahimi-etal-2019-massively} \\
        \midrule
        {\scshape XNLI} & NLI & \makecell[tl]{ar, bg, de, el, en,\\es, ru, sw, ur, zh} & \citet{conneau2018xnli} \\
        IndicXNLI & NLI & ml, te & \citet{aggarwal-etal-2022-indicxnli} \\
        CS ANLI  & NLI & cs, sk* & \citetalias{cs-anli} \\
        RoNLI & NLI & ro & \citet{poesina-etal-2024-novel} \\
        Sl-NLI & NLI & sl & \citet{klemen-etal-2024-si} \\
        \midrule
        MultiClaim & CWCD & \makecell[tl]{ar, bg, cs, de*, el*, en,\\ es, ml*, ro*, ru*, sl*,\\sk, sw*, te*, ur*, zh*} & \makecell[tl]{\citet{pikuliak2023multilingual}\\ \citet{hyben2023bigger}} \\
        \bottomrule
    \end{tabular}
    }
    \caption{The list of datasets used in our experiments. Languages marked with * represent language versions of datasets that are not original but were obtained by translating texts from Czech (CS ANLI) or English (MultiClaim).}
    \label{tab:datasets}
\end{table}

\subsection{Experimental Setup}

\paragraph{Language Representations.}

Language adapters and soft prompts were trained using a \textit{span corruption} objective with different learning rates for training language adapters and soft language prompts, which were identified based on experiments on English data. Detailed parameters are listed in Table~\ref{tab:hyperparameters} in Appendix~\ref{sec:hyperparameters}.

\paragraph{Task Representations.}

In training task representations, we divided the training set into training and validation splits using 15\% of the records for validation, which was done only for datasets that do not include a test set and the original validation split was considered a test set. This is especially the case for the question answering and check-worthy claim detection tasks. Secondly, we preprocessed each dataset by transforming each record from the particular dataset into the text-to-text format employing prompt templates listed in Appendix~\ref{sec:prompts}.

Task representations in all configurations were trained using the same training parameters across all tasks, with differences only between learning rates and weight decay.\footnote{We employed only one seed due to computational and time limitations. However, we performed a check of the generalizability of the approach by training the task representation on the German version of the WikiANN dataset for NER using two additional seeds and evaluated cross-lingual transfer from German to six languages. The results are in Appendix~\ref{app:multiple-seeds}.} In addition, the instruction used for training soft prompts differs across languages and tasks. These variations are based on the language in which the answer is to be generated and the task that the LLM is solving. 

The best model was chosen based on the performance on the validation split with respect to the loss. For classification tasks, we set the maximum number of tokens to generate based on the predicted classes. This minimizes the problem that the LLM continues to generate an answer and enables us to evaluate the LLM's performance correctly. Table~\ref{tab:hyperparameters} in Appendix~\ref{sec:hyperparameters} shows the exact parameters for training language and task representations.

\paragraph{Evaluation.}

For evaluation, we selected several standard metrics employed for particular tasks. Specifically, we use the F1-Score or Accuracy for classification tasks and QA in the {\scshape SQuAD} format. Besides the F1-Score for QA, we also calculated Exact Match, assessing how many of the answers exactly match the ground truth.\footnote{Exact Match tends to underestimate models' performance for low-resource languages, where LLMs are not often able to produce the exact answer with the correct grammar.} For the evaluation, we employed metrics implemented in the Hugging Face evaluate library\footnote{\url{https://huggingface.co/docs/evaluate}}.

We evaluated the results on cross-lingual transfer from high-resource languages to mid- and low-resource ones, where task representations were trained on datasets in high-resource languages. We aim to assess the combination of language representations of low-resource languages with task representations trained on datasets from high-resource languages, i.e., high-resource language as source language and low-resource as target ones. Extended results are shown in Appendix~\ref{sec:all-results}.

\paragraph{Baselines.} 

To evaluate the proposed methods, we employed several baseline approaches and configurations. Baselines include task adapters, soft task prompts (prompt-tuning approach), and MAD-X,  the combination of a language and task adapter~\citet{pfeiffer-etal-2020-mad}. These baselines provided a foundation for assessing the effectiveness of cross-lingual transfer in our experiments.

\section{Results and Analyis}

\paragraph{Overall Results.}

\begin{figure}[t]
\centering
\includegraphics[width=.9\columnwidth]{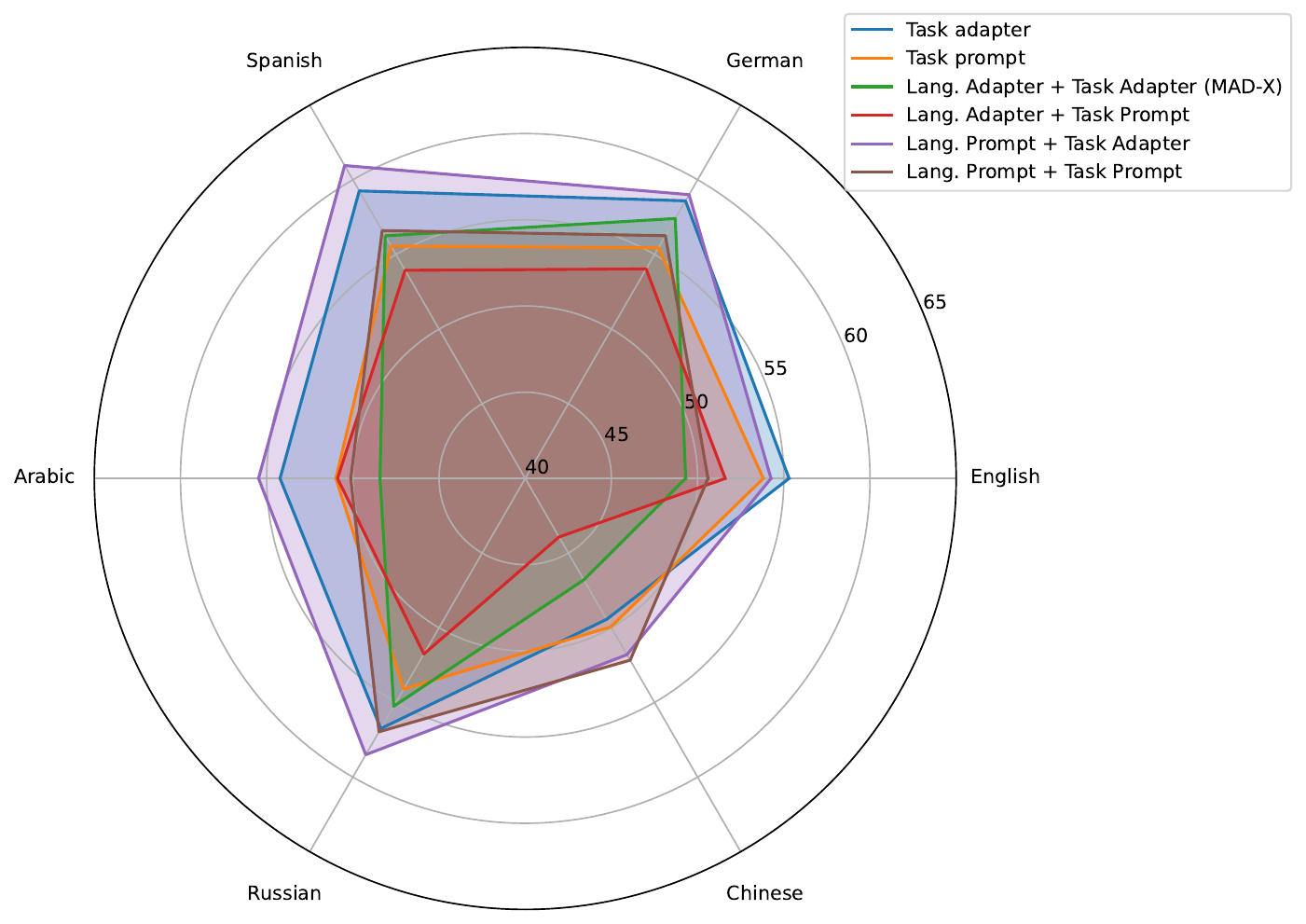}
\caption{Average performance for the transfer from the 6 high-resource languages to all low-resource languages, averaged over all low-resource languages. The graph compares different configurations with varying performance for cross-lingual transfer from high-resource to low-resource languages. In most cases, the combination of soft language prompts with task adapters (purple) proved best.}
\label{fig:radar}
\end{figure}

Our study on cross-lingual transfer performance between high-resource and mid- and low-resource languages is summarized in Table~\ref{tab:summary} in Appendix~\ref{sec:all-results}, which reports averaged metrics across four tasks for mid- and low-resource languages. Additionally, Figure~\ref{fig:radar} demonstrates the comparison of all combinations across high-resource languages, where the presented scores represent the average calculated across all tasks and all mid- and low-resource languages. 

The results demonstrate that the selection of source languages plays an important role in the overall results, with distinct languages demonstrating different performance gains. Using English as a source language yielded the highest performance for most mid- and low-resource languages when employing task representations alone. A possible explanation might be that multilingual models often remain biased toward the source language, even after adaptation, as demonstrated in~\citet{alabi-etal-2024-hidden}. They show that language adaptation in models primarily occurs in the final layers, while earlier predictions are still influenced by the source language. However, for Bulgarian and Slovak, the combination of soft language prompts with task adapters proved to be more effective. 

In contrast, when using Arabic, German, Spanish and Russian as source languages, configurations combining language and task representations yielded superior scores. Specifically, transferring knowledge from Spanish using a combination of soft language prompts and task adapters resulted in the highest performance. Therefore, this configuration using Spanish enhanced the model's performance, making Spanish the most effective high-resource language for cross-lingual transfer between languages across various scripts.

\paragraph{Question Answering.}

Our experiments (see Table~\ref{tab:qa-all}) revealed that the configuration of a soft language prompt and task adapter achieved the highest performance in many cases in the QA task when transferring to mid- and low-resource languages, with only small differences across languages. This configuration was particularly effective for Greek, Romanian and Slovak, while for Telugu and Urdu, the task adapter without language representation outperformed other configurations. This suggests that the complexity of the target language cannot be sufficiently modeled based on the small number of Wikipedia articles in those languages. Furthermore, English excelled across the board, particularly with Latin and Greek scripts, showcasing its adaptability in cross-lingual transfer.

\begin{figure}
\centering
\includegraphics[width=0.75\columnwidth]{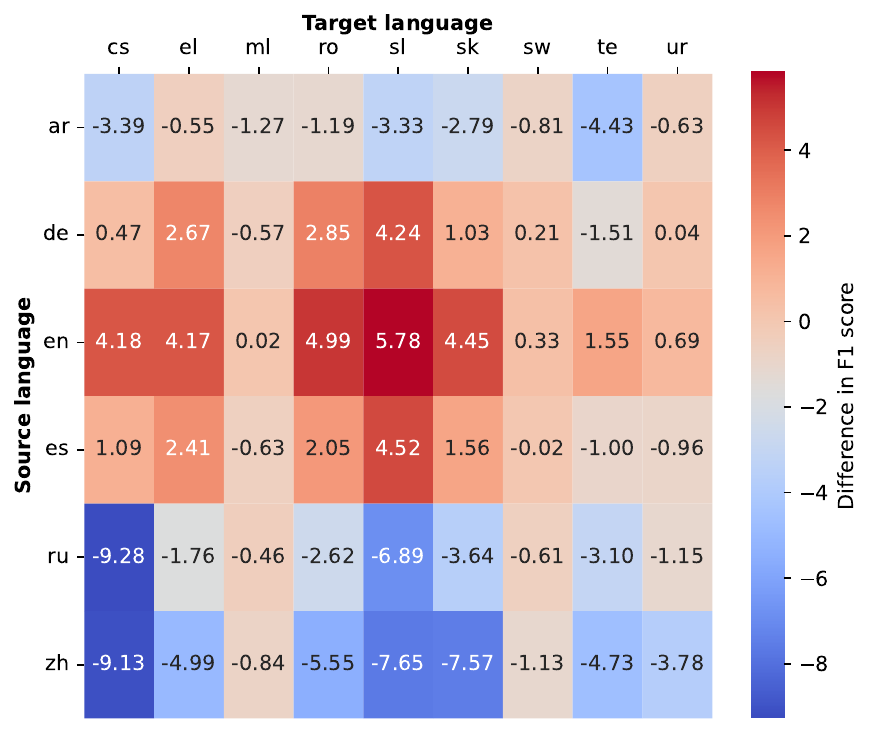}
\caption{Relative F1 improvement for the QA task in transferring knowledge between languages using soft language prompts and task adapters. The effectiveness of the selected configuration is compared with the results obtained without using any language and task representations (i.e., mT0-{\scshape Base} inference).}
\label{fig:qa-heatmap}
\end{figure}

In addition to investigating the effects of individual configurations, we also evaluated the improvement of a soft language prompt combined with a task adapter over the original mT0-{\scshape Base} model without any language or task representations (see Figure~\ref{fig:qa-heatmap}). Figure~\ref{fig:qa-heatmap} contains relative F1-Score improvements and demonstrates that training task representations in English and evaluating in other languages provide the most evident improvement. We also observed that German, English and Spanish improved performance for most low-resource languages, with the exception of Telugu and Malayalam. In contrast, Arabic, Russian and Chinese, which have different scripts, exhibited negative transfer across all cases, with Arabic and Chinese offering no improvement for any languages. We conjecture that the cross-lingual transfer depends on the script used for the language, where we achieved the highest performance for languages in the Latin script. 

\paragraph{Named Entity Recognition.}

In the case of the NER task, Arabic, German, Spanish and Russian, among high-resource languages, performed best in cross-lingual transfer to mid- and low-resource languages, while English and Chinese performed poorly. However, based on the results in Table~\ref{tab:ner-all}, the best improvements were observed using a soft language prompt with a task adapter, outperforming the combination of language and task adapters for languages, such as Arabic, Spanish and German. This is especially the case for Telugu, where the difference between these configurations is more than 37\% in favor of the combination of soft language prompt and task adapter using Russian data.

\paragraph{Natural Language Inference.}

The cross-lingual evaluation of the NLI task from Table~\ref{tab:nli-all} demonstrated the effectiveness of almost all proposed configurations for knowledge transfer. In particular, we mostly achieve superior results using the combination of language adapters with soft task prompts in Czech, Slovenian and Slovak as target languages. While for Swahili, Telugu and Urdu, the best performance was achieved without employing language representations. Furthermore, the high effect on the Romanian language observed in the cross-lingual evaluation is probably because Romanian has been involved during the training of mT5, but not as part of fine-tuning the mT0 model.

Across the six proposed configurations for transferring knowledge, we observed improvement for most languages. However, for Czech, Slovenian and Slovak, several configurations resulted in lower performance compared to inference-only baselines. Notably, for Slovenian, using Russian for the soft task prompt was the only configuration that outperformed the inference-only approach. Furthermore, the combination of language and task adapters for Slovenian resulted in the poorest performance, with an average deterioration of 63\%.

\paragraph{Check-Worthy Claim Detection.}

For check-worthy claim detection, the configuration of soft language prompts and task adapters performs comparably to methods without language representations (see Table~\ref{tab:cwcd-all}). When considering both the best and second-best results, this combination proves effective across most language pairs, demonstrating the model's enhanced capabilities for check-worthy claim detection. Notably, using Spanish for knowledge transfer within this setup resulted in the highest performance gains.

%% file: conclusion.tex
\section{Discussion}

Based on our experiments, we summarize our observations below.

\paragraph{Prompt Tuning Performs Better with Fine-Tuned Models.}

In our preliminary model selection experiments, we found that prompt tuning does not improve the performance of pre-trained LLMs (e.g., mT5) trained only on unlabelled data for downstream tasks. However, prompt-tuning can enhance the performance of already fine-tuned LLMs on any labelled data, even if the specific tasks were not part of the prior fine-tuning. This was confirmed in our experiments with NER and check-worthy claim detection, where fine-tuned LLMs delivered superior results despite no previous task-specific training on these tasks.

\paragraph{Soft Language Prompts with Task Adapters Perform Best in Many Cases.}

Our approach of combining soft language prompts with task adapters demonstrated better performance in many cases, compared to the approach of combined language and task adapters, which has been shown to be very effective in previous work. Specifically, the combination of soft language prompts and task adapters is most effective on the classification tasks, achieving superior results most often. For languages with a different script (e.g., Spanish and Telugu), these differences were over 20\%.

\paragraph{Language Representations are Unable to Capture Linguistic Characteristics Using Small Number of Unlabelled Data.}

Language representations have several limitations that led to configurations without language representations performing consistently better on cross-lingual transfer to highly low-resource languages, such as Telugu, Urdu, and Malayalam. We postulate that the reason is the small number of Wikipedia articles on which the language representations were trained, rendering them unable to adequately capture sufficient linguistic characteristics.

\section{Conclusion}

Our study provides a comprehensive evaluation of various configurations of adapters and soft prompts for cross-lingual transfer in mid- and low-resource languages. With the systematic evaluation of task adapters, soft task prompts, and combinations of language and task representations, we identified configurations that positively affect LLM's performance across different tasks and languages. Our findings demonstrated that the combination of soft language prompts and task adapters emerged as an effective alternative for transferring knowledge between languages. Furthermore, our findings provide valuable insights for the utilization of a combination of PEFT methods for cross-lingual transfer, while highlighting the need to incorporate language-specific knowledge.

\section*{Limitations}

\paragraph{Model Selection.}

Our analysis of the effectiveness of the language and task representations focused on highly multilingual LLMs that include a wider variety of low-resource languages. From this perspective, there is not a vast number of open-source multilingual LLMs with such extensive language coverage as the mT5 or BLOOM model, while having fewer than 1B parameters. We also considered the AYA model~\cite{ustun2024aya}, but due to limited computational resources, it was not feasible to conduct our experiments. Another aspect of the selection was the involvement of only generative models consisting of encoder-decoder or decoder-only architecture.

\paragraph{Other Languages.}

In selecting appropriate languages, we were limited by the languages covered by the mT5 model. To select high-resource languages, we considered languages that are the most extensive in terms of available resources and are in different scripts, e.g., not only Latin script. On the other hand, when selecting mid- and low-resource languages, we also considered the availability of datasets in multiple languages from different language families as well as the availability of datasets in those languages (both human-annotated and machine-translated).

\paragraph{Other Tasks.}

The tasks in our experiments were selected based on the availability of datasets for each selected language and covered multiple areas of the NLP domain, i.e., reading comprehension, fact-checking, and recognizing textual entailment. We mostly considered tasks involved in the instruction fine-tuning of the mT0-model, but we also included tasks that were not originally used to train the mT0-model, e.g., named-entity recognition and check-worthy claim detection.

\section*{Acknowledgements}

This research was partially supported by \textit{DisAI - Improving scientific excellence and creativity in combating disinformation with artificial intelligence and language technologies}, a project funded by Horizon Europe under \href{https://doi.org/10.3030/101079164}{GA No.101079164}, and by the \textit{MIMEDIS}, a project funded by the Slovak Research and Development Agency under GA No. APVV-21-0114. This work was supported by the Ministry of Education, Youth and Sports of the Czech Republic through the e-INFRA CZ (ID:90254).

%% file: appendix.tex
\section{Computational Resources}

For our experiments, we utilized a computational infrastructure consisting of A10 and A40 NVIDIA GPUs, while our experiments ran in parallel on multiple GPUs. In total, our experiments required around 3,200 GPU hours, ensuring model training and validation for cross-lingual transfer.

\section{Prompts Used}
\label{sec:prompts}

For the purpose of the encoder-decoder model, the record from each dataset needs to be transformed into a text-to-text format. To choose an appropriate prompt format, we experimented with all the prompts used in the mT0 paper~\cite{muennighoff2023crosslingual} and with prompts used in the T5 paper~\cite{10.5555/3455716.3455856}. Prompts, which achieved the best performance during inference with the mT0-{\scshape Base} model, were selected for transforming the records into a text-to-text format. In the following paragraphs, there are the prompts for the individual tasks that have been used to convert to text-to-text format.

\subsection{Question Answering}

\noindent 
\textbf{Template:} \texttt{question: \{question\} context: \{context\}}

\subsection{Natural Language Inference}

\noindent 
\textbf{Template:} \texttt{\{premise\} \textbackslash n\textbackslash n Question: Does this imply that "\{hypothesis\}"? Yes, no, or maybe?}

\subsection{Named-Entity Recognition}

\noindent 
\textbf{Template:} \texttt{tag: \{text\}}

\subsection{Check-Worthy Claim Detection}

\noindent 
\textbf{Template:} \texttt{checkworthiness claim: \{claim\}}

\section{Soft Prompt Initialization}
\label{sec:initialization}

This section includes templates for soft prompts used for the initialization for each language and each task. Templates are divided into language and task templates.

\subsection{Language Templates}

To train language representation using a language modeling objective, we employed a specific prompt that varied only based on the language present in the instruction, leaving the rest of the instruction the same.

The template we used for initialization is as follows: \texttt{"Generate the output in \{Language\}:}", where the \texttt{Language} is replaced by the desired language.

\subsection{Task Templates}

The following are initialization prompt templates for each task, where the instruction depends not only on the task but also on the language.

\paragraph{Question Answering.}

For the question answering task, we utilized \texttt{"Answer the question in \{Language\} language:"}, while replacing \texttt{Language} with the desired language.

\paragraph{Natural Language Inference.}

Natural language inference is the task of assessing whether a hypothesis logically follows from the premise. It is defined as a classification with three possible classes: \textit{entailment}, \textit{contradiction} or \textit{neutral}. However, based on the previous work and instruction tuning of the mT0 model, we replaced above mentioned classes with \textit{Yes}, \textit{No} and \textit{Maybe}, based on the used prompt template.

According to the employed classes, we defined an initialization prompt as follows: \texttt{"Select Yes, No or Maybe based on the implication of the premise on the hypothesis in \{Language\}:"}, while \texttt{Language} is replaced by the desired language.

\paragraph{Named-Entity Recognition.}

The named-entity recognition task aims to identify named entities within the input text. While there are many possible categories, the WikiANN dataset focuses only on detecting three categories: location (LOC), person (PER) and organization (ORG). Based on the defined classes, we created the initialization prompt as follows: \texttt{"Identify NER tags (ORG, PER, LOC) in the text in \{Language\}:"}, where \texttt{Language} is substituted with the specific language.

\paragraph{Check-Worthy Claim Detection.}

The latter task includes check-worthy claim detection, which is a binary classification of assessing whether the given claim is worthy of fact-checking or not. As text labels, we used \textit{Not checkworthy} and \textit{Checkworthy}. This is the initialization prompt for the check-worthy claim detection task: \texttt{Determine whether a given claim in \{Language\} is checkworthy:"}, where \texttt{Language} is replaced by the desired language.

\section{Hyperparameters}
\label{sec:hyperparameters}

Table~\ref{tab:hyperparameters} shows hyperparameters used for training language and task representations using adapters and soft prompts.

\begin{table*}[]
\centering
\small
\begin{tabular}{@{}lllll@{}}
\toprule
\multicolumn{1}{c}{\multirow{2}{*}{\textbf{Hyperparameters}}} & \multicolumn{2}{c}{\textbf{Language Modeling}} & \multicolumn{2}{c}{\textbf{Task Modeling}} \\ \cmidrule(l){2-5} 
\multicolumn{1}{c}{} & \textbf{Language Adapter} & \textbf{Soft Language Prompt} & \multicolumn{1}{c}{\textbf{Task Adapter}} & \multicolumn{1}{c}{\textbf{Soft Task Prompt}} \\ \midrule
\textbf{Learning rate} & 5e-5 & 5e-1 & 5e-5 & 5e-1 \\
\textbf{Weight decay} & 0 & 1e-5 & 0 & 1e-5 \\
\textbf{Batch size} & 32 & 32 & 32 & 32 \\
\textbf{No. Training steps} & 100,000 & 100,000 & 50,000 & 50,000 \\
\textbf{Optimizer} & AdamW & Adafactor & AdamW & Adafactor \\
\textbf{Evaluation steps} & 500 & 500 & 1000 & 1000 \\
\textbf{Max input length} & 256 & 256 & 256 & 256 \\
\textbf{Token size of soft prompt} & NaN & 50 & NaN & 50 \\ \bottomrule
\end{tabular}
\caption{Final parameters employed to train language and task representation using adapters and soft prompts.}
\label{tab:hyperparameters}
\end{table*}

\section{Cross-Lingual Evaluation}
\label{sec:all-results}

Tables~\ref{tab:qa-all} to \ref{tab:cwcd-all} present the results for transferring knowledge from all high-resource languages to all mid- and low-resource languages. The first row in each table represents the scores obtained by inference of the original mT0-{\scshape Base} model without additional training of language or task representations.

\begin{table*}[]
\centering
\small
\resizebox{\textwidth}{!}{%
\begin{tabular}{@{}lllllllllllll@{}}
\toprule[.15em]
\textbf{\begin{tabular}[c]{@{}c@{}}Task\\Language\end{tabular}} & \multicolumn{1}{c}{\textbf{\begin{tabular}[c]{@{}c@{}}Language\\Representation\end{tabular}}} & \multicolumn{1}{c}{\textbf{\begin{tabular}[c]{@{}c@{}}Task\\Representation\end{tabular}}} & \multicolumn{1}{c}{\textbf{bg}} & \multicolumn{1}{c}{\textbf{cs}} & \multicolumn{1}{c}{\textbf{el}} & \multicolumn{1}{c}{\textbf{ml}} & \multicolumn{1}{c}{\textbf{ro}} & \multicolumn{1}{c}{\textbf{sl}} & \multicolumn{1}{c}{\textbf{sk}} & \multicolumn{1}{c}{\textbf{sw}} & \multicolumn{1}{c}{\textbf{te}} & \multicolumn{1}{c}{\textbf{ur}} \\ \midrule[.15em]
\multirow{6}{*}{\textbf{ar}} & \multirow{2}{*}{None} & Adapter & 68.03 & \multicolumn{1}{r}{\textbf{48.98}} & \multicolumn{1}{r}{{\ul 65.25}} & \multicolumn{1}{r}{46.74} & \multicolumn{1}{r}{63.44} & \multicolumn{1}{r}{48.74} & \multicolumn{1}{r}{{\ul 48.26}} & \multicolumn{1}{r}{{\ul 50.86}} & \multicolumn{1}{r}{\textbf{53.80}} & \multicolumn{1}{r}{{\ul 48.15}} \\
 &  & Soft Prompt & 64.67 & \multicolumn{1}{r}{43.16} & \multicolumn{1}{r}{62.66} & \multicolumn{1}{r}{47.55} & \multicolumn{1}{r}{{\ul 64.43}} & \multicolumn{1}{r}{44.60} & \multicolumn{1}{r}{40.33} & \multicolumn{1}{r}{47.22} & \multicolumn{1}{r}{51.44} & \multicolumn{1}{r}{43.81} \\ \cmidrule(l){2-13}
 & \multirow{2}{*}{Adapter} & Adapter & 64.82 & \multicolumn{1}{r}{42.56} & \multicolumn{1}{r}{64.59} & \multicolumn{1}{r}{\textbf{49.82}} & \multicolumn{1}{r}{61.73} & \multicolumn{1}{r}{30.44} & \multicolumn{1}{r}{39.92} & \multicolumn{1}{r}{48.99} & \multicolumn{1}{r}{40.36} & \multicolumn{1}{r}{41.03} \\
 &  & Soft Prompt & {\ul 69.90} & \multicolumn{1}{r}{39.84} & \multicolumn{1}{r}{63.36} & \multicolumn{1}{r}{{\ul 48.49}} & \multicolumn{1}{r}{55.78} & \multicolumn{1}{r}{\textbf{55.53}} & \multicolumn{1}{r}{41.28} & \multicolumn{1}{r}{49.52} & \multicolumn{1}{r}{41.45} & \multicolumn{1}{r}{43.75} \\ \cmidrule(l){2-13} 
 & \multirow{2}{*}{Soft Prompt} & Adapter & \textbf{71.23} & \multicolumn{1}{r}{{\ul 48.06}} & \multicolumn{1}{r}{\textbf{67.22}} & \multicolumn{1}{r}{48.22} & \multicolumn{1}{r}{\textbf{65.66}} & \multicolumn{1}{r}{{\ul 51.68}} & \multicolumn{1}{r}{\textbf{48.26}} & \multicolumn{1}{r}{\textbf{51.21}} & \multicolumn{1}{r}{{\ul 53.54}} & \multicolumn{1}{r}{\textbf{49.63}} \\
 &  & Soft Prompt & 66.73 & \multicolumn{1}{r}{41.59} & \multicolumn{1}{r}{62.82} & \multicolumn{1}{r}{47.66} & \multicolumn{1}{r}{58.81} & \multicolumn{1}{r}{44.69} & \multicolumn{1}{r}{40.02} & \multicolumn{1}{r}{44.10} & \multicolumn{1}{r}{50.11} & \multicolumn{1}{r}{44.72} \\ \midrule[.15em]
 \multirow{6}{*}{\textbf{de}} & \multirow{2}{*}{None} & Adapter & 67.76 & 51.72 & {\ul 71.12} & {\ul 51.52} & \textbf{71.55} & 54.88 & 51.27 & \textbf{56.78} & \textbf{57.50} & {\ul 51.82} \\
 &  & Soft Prompt & 65.31 & 47.61 & 69.55 & \textbf{51.77} & {\ul 70.02} & 50.97 & 46.53 & 53.16 & 52.91 & 46.66 \\ \cmidrule(l){2-13} 
 & \multirow{2}{*}{Adapter} & Adapter & \textbf{79.98} & \textbf{53.81} & \textbf{73.48} & 50.39 & 66.34 & 49.69 & \textbf{54.57} & 54.29 & 45.65 & 45.88 \\
 &  & Soft Prompt & {\ul 72.61} & 47.19 & 68.12 & 48.81 & 60.14 & {\ul 56.41} & 46.63 & 51.20 & 44.41 & 44.76 \\ \cmidrule(l){2-13} 
 & \multirow{2}{*}{Soft Prompt} & Adapter & 70.71 & {\ul 51.91} & 70.83 & 50.86 & 69.86 & \textbf{57.76} & {\ul 53.98} & {\ul 55.90} & {\ul 55.90} & \textbf{52.34} \\
 &  & Soft Prompt & 68.98 & 49.02 & 69.83 & 50.98 & 69.18 & 52.18 & 48.36 & 54.50 & 54.50 & 45.04 \\ \midrule[.15em]
\multirow{6}{*}{\textbf{en}} & \multirow{2}{*}{None} & Adapter & {\ul 68.77} & \textbf{49.64} & \textbf{70.37} & {\ul 46.34} & 67.29 & 52.54 & {\ul 47.98} & 49.70 & {\ul 52.13} & 48.29 \\
 &  & Soft Prompt & 64.81 & 41.26 & 68.02 & \textbf{46.60} & 70.58 & 52.32 & 40.19 & \textbf{51.43} & \textbf{52.43} & {\ul 50.48} \\ \cmidrule(l){2-13} 
 & \multirow{2}{*}{Adapter} & Adapter & 60.90 & 46.51 & 65.03 & 41.80 & {\ul 72.02} & 37.96 & 43.30 & 46.98 & 38.47 & 40.09 \\
 &  & Soft Prompt & 64.68 & 38.29 & 65.04 & 42.93 & \textbf{74.50} & \textbf{54.80} & 37.41 & 48.36 & 37.48 & \textbf{52.54} \\ \cmidrule(l){2-13} 
 & \multirow{2}{*}{Soft Prompt} & Adapter & \textbf{68.95} & {\ul 48.37} & {\ul 69.39} & 43.44 & 64.20 & 51.34 & \textbf{50.88} & {\ul 51.05} & 49.16 & 45.78 \\
 &  & Soft Prompt & 56.33 & 47.75 & 65.98 & 43.22 & 58.94 & {\ul 52.64} & 46.48 & 48.96 & 47.12 & 38.62 \\ \midrule[.15em]
\multirow{6}{*}{\textbf{es}} & \multirow{2}{*}{None} & Adapter & 68.65 & {\ul 53.61} & 70.06 & 49.75 & \textbf{73.83} & {\ul 56.23} & 51.82 & \textbf{57.47} & \textbf{57.12} & \textbf{54.01} \\
 &  & Soft Prompt & 62.16 & 50.28 & 66.95 & 47.26 & 71.84 & 50.34 & 49.05 & 54.09 & 53.94 & 49.71 \\ \cmidrule(l){2-13} 
 & \multirow{2}{*}{Adapter} & Adapter & {\ul 73.89} & 50.03 & {\ul 73.33} & {\ul 51.15} & 70.92 & 45.79 & {\ul 53.04} & 54.24 & 45.45 & 44.63 \\
 &  & Soft Prompt & 72.63 & 50.19 & 64.62 & 44.62 & 72.70 & 47.07 & 51.71 & 51.95 & 39.95 & 43.94 \\ \cmidrule(l){2-13} 
 & \multirow{2}{*}{Soft Prompt} & Adapter & \textbf{75.37} & \textbf{55.23} & \textbf{73.50} & \textbf{51.39} & 72.03 & \textbf{58.88} & \textbf{54.84} & {\ul 57.41} & {\ul 56.90} & {\ul 54.00} \\
 &  & Soft Prompt & 69.07 & 52.16 & 64.43 & 48.19 & {\ul 72.98} & 53.21 & 50.93 & 53.87 & 50.89 & 50.26 \\ \midrule[.15em]
\multirow{6}{*}{\textbf{ru}} & \multirow{2}{*}{None} & Adapter & {\ul 82.44} & \multicolumn{1}{r}{{\ul 45.29}} & \multicolumn{1}{r}{65.74} & \multicolumn{1}{r}{46.79} & \multicolumn{1}{r}{69.68} & \multicolumn{1}{r}{49.24} & \multicolumn{1}{r}{45.24} & \multicolumn{1}{r}{{\ul 54.40}} & \multicolumn{1}{r}{{\ul 56.48}} & \multicolumn{1}{r}{{\ul 52.27}} \\
 &  & Soft Prompt & 79.26 & \multicolumn{1}{r}{41.51} & \multicolumn{1}{r}{60.83} & \multicolumn{1}{r}{48.64} & \multicolumn{1}{r}{68.09} & \multicolumn{1}{r}{45.29} & \multicolumn{1}{r}{40.70} & \multicolumn{1}{r}{50.73} & \multicolumn{1}{r}{55.13} & \multicolumn{1}{r}{50.94} \\ \cmidrule(l){2-13} 
 & \multirow{2}{*}{Adapter} & Adapter & 80.74 & \multicolumn{1}{r}{\textbf{52.52}} & \multicolumn{1}{r}{\textbf{73.86}} & \multicolumn{1}{r}{45.76} & \multicolumn{1}{r}{\textbf{73.75}} & \multicolumn{1}{r}{40.19} & \multicolumn{1}{r}{\textbf{51.99}} & \multicolumn{1}{r}{52.08} & \multicolumn{1}{r}{39.55} & \multicolumn{1}{r}{42.11} \\
 &  & Soft Prompt & 77.96 & \multicolumn{1}{r}{34.33} & \multicolumn{1}{r}{66.75} & \multicolumn{1}{r}{44.48} & \multicolumn{1}{r}{71.76} & \multicolumn{1}{r}{49.84} & \multicolumn{1}{r}{36.41} & \multicolumn{1}{r}{50.61} & \multicolumn{1}{r}{40.14} & \multicolumn{1}{r}{45.31} \\ \cmidrule(l){2-13} 
 & \multirow{2}{*}{Soft Prompt} & Adapter & \textbf{83.96} & \multicolumn{1}{r}{44.69} & \multicolumn{1}{r}{70.25} & \multicolumn{1}{r}{{\ul 49.87}} & \multicolumn{1}{r}{{\ul 72.15}} & \multicolumn{1}{r}{{\ul 51.06}} & \multicolumn{1}{r}{{\ul 47.34}} & \multicolumn{1}{r}{\textbf{55.21}} & \multicolumn{1}{r}{\textbf{56.89}} & \multicolumn{1}{r}{\textbf{53.62}} \\
 &  & Soft Prompt & 79.68 & \multicolumn{1}{r}{44.50} & \multicolumn{1}{r}{{\ul 71.02}} & \multicolumn{1}{r}{\textbf{50.63}} & \multicolumn{1}{r}{70.94} & \multicolumn{1}{r}{\textbf{52.60}} & \multicolumn{1}{r}{40.99} & \multicolumn{1}{r}{52.37} & \multicolumn{1}{r}{54.91} & \multicolumn{1}{r}{52.04} \\ \midrule[.15em]
\multirow{6}{*}{\textbf{zh}} & \multirow{2}{*}{None} & Adapter & 61.19 & \multicolumn{1}{r}{43.74} & \multicolumn{1}{r}{57.03} & \multicolumn{1}{r}{40.50} & \multicolumn{1}{r}{64.60} & \multicolumn{1}{r}{44.42} & \multicolumn{1}{r}{42.70} & \multicolumn{1}{r}{{\ul 50.23}} & \multicolumn{1}{r}{45.65} & \multicolumn{1}{r}{{\ul 44.33}} \\
 &  & Soft Prompt & 58.20 & \multicolumn{1}{r}{45.17} & \multicolumn{1}{r}{59.24} & \multicolumn{1}{r}{{\ul 42.23}} & \multicolumn{1}{r}{{\ul 66.36}} & \multicolumn{1}{r}{44.72} & \multicolumn{1}{r}{43.38} & \multicolumn{1}{r}{46.27} & \multicolumn{1}{r}{50.10} & \multicolumn{1}{r}{43.72} \\ \cmidrule(l){2-13} 
 & \multirow{2}{*}{Adapter} & Adapter & {\ul 62.73} & \multicolumn{1}{r}{{\ul 46.45}} & \multicolumn{1}{r}{{\ul 61.14}} & \multicolumn{1}{r}{37.45} & \multicolumn{1}{r}{66.19} & \multicolumn{1}{r}{35.30} & \multicolumn{1}{r}{42.70} & \multicolumn{1}{r}{43.73} & \multicolumn{1}{r}{35.98} & \multicolumn{1}{r}{36.28} \\
 &  & Soft Prompt & 49.40 & \multicolumn{1}{r}{43.50} & \multicolumn{1}{r}{47.04} & \multicolumn{1}{r}{42.18} & \multicolumn{1}{r}{45.70} & \multicolumn{1}{r}{\textbf{56.09}} & \multicolumn{1}{r}{36.34} & \multicolumn{1}{r}{42.75} & \multicolumn{1}{r}{37.22} & \multicolumn{1}{r}{39.00} \\ \cmidrule(l){2-13} 
 & \multirow{2}{*}{Soft Prompt} & Adapter & \textbf{65.97} & \multicolumn{1}{r}{45.20} & \multicolumn{1}{r}{60.49} & \multicolumn{1}{r}{41.62} & \multicolumn{1}{r}{65.64} & \multicolumn{1}{r}{48.49} & \multicolumn{1}{r}{\textbf{45.52}} & \multicolumn{1}{r}{\textbf{52.19}} & \multicolumn{1}{r}{{\ul 50.20}} & \multicolumn{1}{r}{42.72} \\
 &  & Soft Prompt & 62.44 & \multicolumn{1}{r}{\textbf{48.74}} & \multicolumn{1}{r}{\textbf{61.62}} & \multicolumn{1}{r}{\textbf{43.09}} & \multicolumn{1}{r}{\textbf{67.48}} & \multicolumn{1}{r}{{\ul 49.20}} & \multicolumn{1}{r}{{\ul 44.85}} & \multicolumn{1}{r}{48.53} & \multicolumn{1}{r}{\textbf{51.09}} & \multicolumn{1}{r}{\textbf{44.69}} \\ \bottomrule[.15em]
\end{tabular}%
}
\caption{Average scores for each configuration across all tasks for low-resource languages. The languages in rows represent the language in which the task representation was trained, and the languages in columns represent the language representation that was used, if any (except for configurations with None in the language representation). For each language pair, the best results are \textbf{boldfaced} and the second best are {\ul underlined}.}
\label{tab:summary}
\end{table*}

\begin{table*}[]
\centering
\resizebox{\textwidth}{!}{%
\begin{tabular}{crrrrrrrrrrrr}
\toprule[.15em] 
\multicolumn{1}{c}{\textbf{\begin{tabular}[c]{@{}c@{}}Task\\Language\end{tabular}}} & \multicolumn{1}{c}{\textbf{\begin{tabular}[c]{@{}c@{}}Language\\ Representation\end{tabular}}} & \multicolumn{1}{c}{\textbf{\begin{tabular}[c]{@{}c@{}}Task\\ Representation\end{tabular}}} & \multicolumn{1}{c}{\textbf{cs}} & \multicolumn{1}{c}{\textbf{el}} & \multicolumn{1}{c}{\textbf{ml}} & \multicolumn{1}{c}{\textbf{ro}} & \multicolumn{1}{c}{\textbf{sl}} & \multicolumn{1}{c}{\textbf{sk}} & \multicolumn{1}{c}{\textbf{sw}} & \multicolumn{1}{c}{\textbf{te}} & \multicolumn{1}{c}{\textbf{ur}} \\ \midrule[.15em]
\textbf{} & None & None & 31.34 (24.78) & 57.00 (47.56) & 1.37 (1.07) & 57.00 (47.56) & 31.50 (22.58) & 26.39 (9.78) & 3.24 (0.36) & 18.64 (12.10) & 13.37 (7.02) \\ \midrule[.15em]
 &  & Adapter & {\color[HTML]{FE0000} \textbf{29.14 (21.80)}} & {\color[HTML]{FE0000} 55.90 (46.30)} & {\color[HTML]{FE0000} 0.22 (18.94)} & {\color[HTML]{FE0000} 55.90 (46.30)} & {\color[HTML]{FE0000} {\ul 27.83 (19.74)}} & {\color[HTML]{FE0000} 23.15 (8.44)} & {\color[HTML]{FE0000} 2.39 (0.36)} & {\color[HTML]{FE0000} \textbf{15.70 (11.00)}} & {\color[HTML]{FE0000} \textbf{13.00 (10.38)}} \\
 & \multirow{-2}{*}{None} & Soft Prompt & {\color[HTML]{FE0000} 22.93 (18.02)} & {\color[HTML]{FE0000} {\ul 56.08 (46.89)}} & {\color[HTML]{FE0000} 0.10 (0.82)} & {\color[HTML]{FE0000} {\ul 56.08 (46.89)}} & {\color[HTML]{FE0000} 23.91 (17.33)} & {\color[HTML]{FE0000} 19.89 (7.85)} & {\color[HTML]{FE0000} 0.56 (0.18)} & {\color[HTML]{FE0000} 11.53 (8.60)} & {\color[HTML]{FE0000} 12.05 (8.29)} \\ \cmidrule(l){2-12} 
 &  & Adapter & {\color[HTML]{FE0000} 25.60 (14.45)} & {\color[HTML]{FE0000} 45.96 (35.71)} & {\color[HTML]{FE0000} \textbf{1.12 (4.15)}} & {\color[HTML]{FE0000} 43.17 (31.85)} & {\color[HTML]{FE0000} 23.55 (11.82)} & {\color[HTML]{FE0000} 22.84 (6.03)} & {\color[HTML]{FE0000} 1.77 (0.18)} & {\color[HTML]{FE0000} 9.54 (4.90)} & {\color[HTML]{FE0000} 9.04 (4.08)} \\
 & \multirow{-2}{*}{Adapter} & Soft Prompt & {\color[HTML]{FE0000} 24.70 (1.67)} & {\color[HTML]{FE0000} 52.76 (42.94)} & {\color[HTML]{FE0000} {\ul 0.95 (0.63)}} & {\color[HTML]{FE0000} 46.35 (35.38)} & {\color[HTML]{FE0000} 27.65 (16.04)} & {\color[HTML]{FE0000} \textbf{25.10 (1.28)}} & {\color[HTML]{FE0000} \textbf{2.57 (1.09)}} & {\color[HTML]{FE0000} 10.50 (5.10)} & {\color[HTML]{FE0000} 10.43 (5.84)} \\ \cmidrule(l){2-12}
 &  & Adapter & {\color[HTML]{FE0000} {\ul 27.95 (20.78)}} & {\color[HTML]{FE0000} \textbf{56.45 (46.97)}} & {\color[HTML]{FE0000} 0.10 (21.27)} & {\color[HTML]{FE0000} 55.81 (46.47)} & {\color[HTML]{FE0000} \textbf{28.17 (19.88)}} & {\color[HTML]{FE0000} {\ul 23.60 (8.72)}} & {\color[HTML]{FE0000} {\ul 2.43 (0.45)}} & {\color[HTML]{FE0000} {\ul 14.21 (10.10)}} & {\color[HTML]{FE0000} {\ul 12.74 (10.07)}} \\
\multirow{-6}{*}{\textbf{ar}} & \multirow{-2}{*}{Soft Prompt} & Soft Prompt & {\color[HTML]{FE0000} 27.14 (20.55)} & {\color[HTML]{FE0000} 55.51 (46.72)} & {\color[HTML]{FE0000} 0.03 (0.31)} & {\color[HTML]{FE0000} \textbf{56.11 (47.23)}} & {\color[HTML]{FE0000} 26.00 (18.62)} & {\color[HTML]{FE0000} 20.96 (8.04)} & {\color[HTML]{FE0000} 0.77 (0.27)} & {\color[HTML]{FE0000} 11.04 (8.30)} & {\color[HTML]{FE0000} 12.37 (8.63)} \\ \midrule[.15em]
 &  & Adapter & {\color[HTML]{009901} 35.37 (27.29)} & {\color[HTML]{009901} {\ul 58.88 (49.08)}} & {\color[HTML]{FE0000} 0.99 (2.27)} & {\color[HTML]{009901} {\ul 58.88 (49.08)}} & {\color[HTML]{009901} 36.15 (24.66)} & {\color[HTML]{009901} 27.51 (10.16)} & {\color[HTML]{009901} \textbf{3.84 (0.82)}} & {\color[HTML]{009901} \textbf{18.81 (12.80)}} & {\color[HTML]{009901} \textbf{14.02 (10.27)}} \\
 & \multirow{-2}{*}{None} & Soft Prompt & {\color[HTML]{FE0000} 28.56 (21.31)} & {\color[HTML]{009901} 57.12 (47.65)} & {\color[HTML]{FE0000} 0.57 (2.20)} & {\color[HTML]{009901} 57.12 (47.65)} & {\color[HTML]{FE0000} 30.34 (20.52)} & {\color[HTML]{FE0000} 24.54 (8.92)} & {\color[HTML]{FE0000} 1.93 (0.73)} & {\color[HTML]{FE0000} 12.46 (9.70)} & {\color[HTML]{FE0000} 10.54 (8.21)} \\ \cmidrule(l){2-12} 
 &  & Adapter & {\color[HTML]{009901} {\ul 36.78 (27.82)}} & {\color[HTML]{009901} 57.46 (47.31)} & {\color[HTML]{FE0000} {\ul 1.05 (1.38)}} & {\color[HTML]{009901} 57.69 (47.90)} & {\color[HTML]{009901} {\ul 38.03 (25.07)}} & {\color[HTML]{009901} \textbf{31.43 (11.35)}} & {\color[HTML]{FE0000} 3.01 (0.54)} & {\color[HTML]{FE0000} 11.32 (6.00)} & {\color[HTML]{FE0000} 9.62 (4.72)} \\
 & \multirow{-2}{*}{Adapter} & Soft Prompt & {\color[HTML]{009901} \textbf{38.13 (31.20)}} & {\color[HTML]{FE0000} 54.70 (45.29)} & {\color[HTML]{009901} \textbf{1.88 (0.94)}} & {\color[HTML]{FE0000} 55.11 (46.55)} & {\color[HTML]{009901} \textbf{38.88 (27.69)}} & {\color[HTML]{009901} {\ul 30.70 (11.98)}} & {\color[HTML]{FE0000} 3.13 (1.00)} & {\color[HTML]{FE0000} 16.05 (8.70)} & {\color[HTML]{FE0000} 12.63 (8.01)} \\ \cmidrule(l){2-12} 
 &  & Adapter & {\color[HTML]{009901} 31.81 (24.29)} & {\color[HTML]{009901} \textbf{59.67 (48.91)}} & {\color[HTML]{FE0000} 0.80 (2.08)} & {\color[HTML]{009901} \textbf{59.85 (49.08)}} & {\color[HTML]{009901} 35.74 (23.84)} & {\color[HTML]{009901} 27.42 (10.16)} & {\color[HTML]{009901} {\ul 3.45 (0.36)}} & {\color[HTML]{FE0000} {\ul 17.13 (11.20)}} & {\color[HTML]{009901} {\ul 13.41 (10.25)}} \\
\multirow{-6}{*}{\textbf{de}} & \multirow{-2}{*}{Soft Prompt} & Soft Prompt & {\color[HTML]{009901} 32.68 (24.65)} & {\color[HTML]{009901} 57.86 (48.15)} & {\color[HTML]{FE0000} 0.58 (8.68)} & {\color[HTML]{009901} 58.38 (49.08)} & {\color[HTML]{FE0000} 30.67 (20.78)} & {\color[HTML]{009901} 27.95 (10.30)} & {\color[HTML]{FE0000} 2.25 (1.09)} & {\color[HTML]{FE0000} 12.34 (9.50)} & {\color[HTML]{FE0000} 9.35 (9.58)} \\\midrule[.15em]
 &  & Adapter & {\color[HTML]{009901} \textbf{36.95 (28.57)}} & {\color[HTML]{009901} 60.24 (50.34)} & {\color[HTML]{FE0000} 1.18 (0.94)} & {\color[HTML]{009901} 60.24 (50.34)} & {\color[HTML]{009901} {\ul 37.04 (26.07)}} & {\color[HTML]{009901} {\ul 30.11 (11.46)}} & {\color[HTML]{FE0000} 3.21 (0.36)} & {\color[HTML]{009901} {\ul 19.65 (12.70)}} & {\color[HTML]{009901} {\ul 13.93 (9.37)}} \\
 & \multirow{-2}{*}{None} & Soft Prompt & {\color[HTML]{009901} 33.59 (25.55)} & {\color[HTML]{009901} {\ul 60.35 (50.76)}} & {\color[HTML]{FE0000} 0.82 (0.94)} & {\color[HTML]{009901} {\ul 60.35 (50.76)}} & {\color[HTML]{009901} 34.77 (24.36)} & {\color[HTML]{009901} 27.76 (10.07)} & {\color[HTML]{FE0000} 2.81 (0.45)} & {\color[HTML]{009901} 19.38 (12.80)} & {\color[HTML]{009901} 13.81 (9.25)} \\ \cmidrule(l){2-12}
 &  & Adapter & {\color[HTML]{009901} 33.75 (24.33)} & {\color[HTML]{009901} 57.44 (48.24)} & {\color[HTML]{FE0000} {\ul 1.25 (1.70)}} & {\color[HTML]{009901} 58.11 (49.33)} & {\color[HTML]{009901} 31.65 (19.76)} & {\color[HTML]{009901} 28.98 (10.58)} & {\color[HTML]{009901} 3.25 (0.63)} & {\color[HTML]{FE0000} 9.85 (5.00)} & {\color[HTML]{FE0000} 9.06 (4.20)} \\
 & \multirow{-2}{*}{Adapter} & Soft Prompt & {\color[HTML]{009901} 33.94 (26.22)} & {\color[HTML]{009901} 58.49 (49.58)} & {\color[HTML]{FE0000} 0.97 (0.25)} & {\color[HTML]{009901} 57.72 (48.74)} & {\color[HTML]{009901} 35.89 (24.36)} & {\color[HTML]{009901} 29.82 (11.54)} & {\color[HTML]{FE0000} 3.06 (0.45)} & {\color[HTML]{FE0000} 12.67 (6.60)} & {\color[HTML]{FE0000} 9.67 (4.94)} \\ \cmidrule(l){2-12}
 &  & Adapter & {\color[HTML]{009901} {\ul 35.52 (27.27)}} & {\color[HTML]{009901} \textbf{61.17 (51.68)}} & {\color[HTML]{009901} \textbf{1.39 (2.71)}} & {\color[HTML]{009901} \textbf{61.99 (52.10)}} & {\color[HTML]{009901} \textbf{37.28 (25.96)}} & {\color[HTML]{009901} \textbf{30.84 (11.73)}} & {\color[HTML]{009901} \textbf{3.57 (0.27)}} & {\color[HTML]{009901} \textbf{20.19 (13.50)}} & {\color[HTML]{009901} \textbf{14.06 (10.54)}} \\
\multirow{-6}{*}{\textbf{en}} & \multirow{-2}{*}{Soft Prompt} & Soft Prompt & {\color[HTML]{009901} 35.35 (27.02)} & {\color[HTML]{009901} 59.80 (50.34)} & {\color[HTML]{FE0000} 0.68 (2.20)} & {\color[HTML]{009901} 60.02 (50.34)} & {\color[HTML]{009901} 36.49 (26.02)} & {\color[HTML]{009901} 29.89 (11.39)} & {\color[HTML]{009901} {\ul 3.46 (0.54)}} & {\color[HTML]{FE0000} 11.71 (7.30)} & {\color[HTML]{FE0000} 9.83 (5.68)} \\ \midrule[.15em]
 &  & Adapter & {\color[HTML]{009901} {\ul 33.72 (25.43)}} & {\color[HTML]{009901} \textbf{59.61 (50.34)}} & {\color[HTML]{FE0000} \textbf{1.29 (1.20)}} & {\color[HTML]{009901} \textbf{59.61 (50.34)}} & {\color[HTML]{009901} {\ul 34.97 (32.80)}} & {\color[HTML]{009901} 27.24 (9.71)} & {\color[HTML]{009901} \textbf{3.42 (0.82)}} & {\color[HTML]{009901} \textbf{19.10 (12.50)}} & {\color[HTML]{FE0000} \textbf{13.03 (9.48)}} \\
 & \multirow{-2}{*}{None} & Soft Prompt & {\color[HTML]{FE0000} 25.98 (19.06)} & {\color[HTML]{FE0000} 54.72 (45.29)} & {\color[HTML]{FE0000} 0.12 (0.31)} & {\color[HTML]{FE0000} 54.72 (45.29)} & {\color[HTML]{FE0000} 27.38 (18.50)} & {\color[HTML]{FE0000} 22.43 (7.62)} & {\color[HTML]{FE0000} 1.10 (0.36)} & {\color[HTML]{FE0000} 11.94 (9.00)} & {\color[HTML]{FE0000} 9.13 (6.84)} \\ \cmidrule(l){2-12}
 &  & Adapter & {\color[HTML]{009901} \textbf{33.98 (23.88)}} & {\color[HTML]{FE0000} 55.44 (45.13)} & {\color[HTML]{FE0000} {\ul 1.06 (1.26)}} & {\color[HTML]{FE0000} 56.58 (46.22)} & {\color[HTML]{009901} 34.52 (21.06)} & {\color[HTML]{009901} 28.66 (10.01)} & {\color[HTML]{FE0000} 2.34 (0.45)} & {\color[HTML]{FE0000} 10.76 (5.00)} & {\color[HTML]{FE0000} 9.12 (4.81)} \\
 & \multirow{-2}{*}{Adapter} & Soft Prompt & {\color[HTML]{009901} 32.86 (24.45)} & {\color[HTML]{FE0000} 53.57 (43.78)} & {\color[HTML]{FE0000} 0.98 (0.57)} & {\color[HTML]{FE0000} 52.82 (43.45)} & {\color[HTML]{009901} 34.73 (23.35)} & {\color[HTML]{009901} {\ul 27.82 (10.35)}} & {\color[HTML]{FE0000} 2.49 (0.54)} & {\color[HTML]{FE0000} 10.89 (5.40)} & {\color[HTML]{FE0000} 9.27 (4.17)} \\ \cmidrule(l){2-12}
 &  & Adapter & {\color[HTML]{009901} 32.43 (24.92)} & {\color[HTML]{009901} {\ul 59.41 (49.75)}} & {\color[HTML]{FE0000} 0.74 (1.07)} & {\color[HTML]{009901} {\ul 59.05 (49.24)}} & {\color[HTML]{009901} \textbf{36.02 (24.92)}} & {\color[HTML]{009901} \textbf{27.95 (10.27)}} & {\color[HTML]{FE0000} {\ul 3.22 (0.45)}} & {\color[HTML]{FE0000} {\ul 17.64 (11.40)}} & {\color[HTML]{FE0000} {\ul 12.41 (8.42)}} \\
\multirow{-6}{*}{\textbf{es}} & \multirow{-2}{*}{Soft Prompt} & Soft Prompt & {\color[HTML]{FE0000} 27.21 (19.82)} & {\color[HTML]{FE0000} 54.39 (44.45)} & {\color[HTML]{FE0000} 0.27 (0.82)} & {\color[HTML]{FE0000} 54.93 (45.21)} & {\color[HTML]{FE0000} 28.25 (19.39)} & {\color[HTML]{FE0000} 21.49 (7.49)} & {\color[HTML]{FE0000} 1.11 (0.18)} & {\color[HTML]{FE0000} 9.28 (7.20)} & {\color[HTML]{FE0000} 8.36 (6.11)} \\ \midrule[.15em]
 &  & Adapter & {\color[HTML]{FE0000} 27.45 (15.86)} & {\color[HTML]{FE0000} \textbf{55.56 (42.52)}} & {\color[HTML]{FE0000} 0.73 (5.73)} & {\color[HTML]{FE0000} \textbf{55.56 (42.52)}} & {\color[HTML]{FE0000} 24.81 (14.05)} & {\color[HTML]{FE0000} 23.69 (7.96)} & {\color[HTML]{FE0000} \textbf{2.93 (0.45)}} & {\color[HTML]{FE0000} {\ul 16.80 (10.50)}} & {\color[HTML]{FE0000} \textbf{12.45 (10.12)}} \\
 & \multirow{-2}{*}{None} & Soft Prompt & {\color[HTML]{FE0000} 17.94 (8.27)} & {\color[HTML]{FE0000} 51.61 (37.73)} & {\color[HTML]{FE0000} 0.25 (0.13)} & {\color[HTML]{FE0000} 51.61 (37.73)} & {\color[HTML]{FE0000} 17.58 (8.13)} & {\color[HTML]{FE0000} 16.98 (4.10)} & {\color[HTML]{FE0000} 0.78 (0.27)} & {\color[HTML]{FE0000} 12.40 (8.00)} & {\color[HTML]{FE0000} 8.71 (5.49)} \\ \cmidrule(l){2-12}
 &  & Adapter & {\color[HTML]{FE0000} 31.24 (14.98)} & {\color[HTML]{FE0000} 54.17 (40.25)} & {\color[HTML]{FE0000} \textbf{1.27 (0.76)}} & {\color[HTML]{FE0000} {\ul 54.72 (40.50)}} & {\color[HTML]{009901} {\ul 32.90 (17.04)}} & {\color[HTML]{009901} {\ul 32.13 (9.71)}} & {\color[HTML]{FE0000} {\ul 2.72 (0.18)}} & {\color[HTML]{FE0000} 10.66 (5.30)} & {\color[HTML]{FE0000} 8.98 (3.85)} \\
 & \multirow{-2}{*}{Adapter} & Soft Prompt & {\color[HTML]{009901} \textbf{32.53 (19.94)}} & {\color[HTML]{FE0000} 51.63 (38.91)} & {\color[HTML]{FE0000} 0.89 (0.25)} & {\color[HTML]{FE0000} 51.78 (37.31)} & {\color[HTML]{009901} \textbf{33.57 (18.08)}} & {\color[HTML]{009901} \textbf{34.00 (13.34)}} & {\color[HTML]{FE0000} 1.96 (0.09)} & {\color[HTML]{FE0000} 10.37 (4.80)} & {\color[HTML]{FE0000} 8.78 (3.91)} \\ \cmidrule(l){2-12}
 &  & Adapter & {\color[HTML]{FE0000} 22.06 (12.33)} & {\color[HTML]{FE0000} {\ul 55.24 (42.44)}} & {\color[HTML]{FE0000} 0.91 (1.32)} & {\color[HTML]{FE0000} 54.38 (41.60)} & {\color[HTML]{FE0000} 24.61 (14.26)} & {\color[HTML]{FE0000} 22.75 (7.53)} & {\color[HTML]{FE0000} 2.63 (0.27)} & {\color[HTML]{FE0000} 15.54 (9.50)} & {\color[HTML]{FE0000} {\ul 12.22 (7.80)}} \\
\multirow{-6}{*}{\textbf{ru}} & \multirow{-2}{*}{Soft Prompt} & Soft Prompt & {\color[HTML]{009901} {\ul 32.13 (18.00)}} & {\color[HTML]{FE0000} 53.90 (40.00)} & {\color[HTML]{FE0000} {\ul 0.99 (0.25)}} & {\color[HTML]{FE0000} 53.37 (39.83)} & {\color[HTML]{FE0000} 29.68 (14.91)} & {\color[HTML]{009901} 31.42 (10.26)} & {\color[HTML]{FE0000} 2.06 (0.18)} & {\color[HTML]{FE0000} \textbf{16.86 (9.20)}} & {\color[HTML]{FE0000} 11.63 (6.04)} \\ \midrule[.15em]
 &  & Adapter & {\color[HTML]{FE0000} 22.06 (16.33)} & {\color[HTML]{FE0000} 50.83 (40.08)} & {\color[HTML]{FE0000} 0.65 (0.50)} & {\color[HTML]{FE0000} 50.83 (40.08)} & {\color[HTML]{FE0000} 21.96 (14.96)} & {\color[HTML]{FE0000} 18.38 (6.74)} & {\color[HTML]{FE0000} 1.42 (0.27)} & {\color[HTML]{FE0000} 13.29 (9.20)} & {\color[HTML]{FE0000} 9.66 (6.31)} \\
 & \multirow{-2}{*}{None} & Soft Prompt & {\color[HTML]{FE0000} {\ul 26.25 (20.33)}} & {\color[HTML]{FE0000} \textbf{56.57 (47.39)}} & {\color[HTML]{FE0000} 0.65 (0.44)} & {\color[HTML]{FE0000} \textbf{56.57 (47.39)}} & {\color[HTML]{FE0000} 25.64 (18.29)} & {\color[HTML]{FE0000} 22.32 (8.53)} & {\color[HTML]{FE0000} 0.91 (0.45)} & {\color[HTML]{FE0000} {\ul 16.00 (11.80)}} & {\color[HTML]{FE0000} {\ul 12.49 (9.17)}} \\ \cmidrule(l){2-12}
 &  & Adapter & {\color[HTML]{FE0000} 26.10 (16.29)} & {\color[HTML]{FE0000} 43.57 (30.34)} & {\color[HTML]{FE0000} \textbf{1.30 (0.57)}} & {\color[HTML]{FE0000} 41.48 (28.57)} & {\color[HTML]{FE0000} {\ul 26.16 (14.21)}} & {\color[HTML]{FE0000} 22.50 (6.17)} & {\color[HTML]{FE0000} 1.94 (0.36)} & {\color[HTML]{FE0000} 11.77 (6.50)} & {\color[HTML]{FE0000} 9.04 (5.01)} \\
 & \multirow{-2}{*}{Adapter} & Soft Prompt & {\color[HTML]{FE0000} 24.83 (14.06)} & {\color[HTML]{FE0000} 47.06 (35.80)} & {\color[HTML]{FE0000} 1.08 (0.06)} & {\color[HTML]{FE0000} 39.35 (25.55)} & {\color[HTML]{FE0000} 25.58 (12.45)} & {\color[HTML]{FE0000} {\ul 24.19 (5.44)}} & {\color[HTML]{FE0000} \textbf{2.64 (0.82)}} & {\color[HTML]{FE0000} 10.34 (4.90)} & {\color[HTML]{FE0000} 8.90 (4.24)} \\ \cmidrule(l){2-12}
 &  & Adapter & {\color[HTML]{FE0000} 22.21 (16.90)} & {\color[HTML]{FE0000} 52.01 (41.01)} & {\color[HTML]{FE0000} 0.53 (0.44)} & {\color[HTML]{FE0000} 51.45 (40.25)} & {\color[HTML]{FE0000} 23.85 (16.20)} & {\color[HTML]{FE0000} 18.82 (6.95)} & {\color[HTML]{FE0000} {\ul 2.11 (0.45)}} & {\color[HTML]{FE0000} 13.91 (9.10)} & {\color[HTML]{FE0000} 9.59 (6.26)} \\
\multirow{-6}{*}{\textbf{zh}} & \multirow{-2}{*}{Soft Prompt} & Soft Prompt & {\color[HTML]{FE0000} \textbf{31.34 (24.20)}} & {\color[HTML]{FE0000} {\ul 56.51 (47.90)}} & {\color[HTML]{FE0000} {\ul 1.23 (0.82)}} & {\color[HTML]{FE0000} {\ul 56.39 (47.98)}} & {\color[HTML]{FE0000} \textbf{29.25 (21.17)}} & {\color[HTML]{FE0000} \textbf{25.06 (9.42)}} & {\color[HTML]{FE0000} 1.85 (0.45)} & {\color[HTML]{FE0000} \textbf{17.52 (12.20)}} & {\color[HTML]{FE0000} \textbf{12.80 (9.48)}} \\ \bottomrule[.15em] 
\end{tabular}}
\caption{Results for the question answering task for cross-lingual transfer from high-resource to mid- and low-resource languages. The results are reported as \textit{F1-Score (Exact Match)}. For each source-target language pair, the best-performing result is highlighted in \textbf{bold}, while the second-best scores are {\ul underlined}. Additionally, language pairs with improved performance compared to inference-only (without incorporating any language or task representation) are marked in {\color[HTML]{009901} green}, and those with decreased performance are marked in {\color[HTML]{FE0000} red}.}
\label{tab:qa-all}
\end{table*}

\begin{table*}[]
\centering
\small
\resizebox{\textwidth}{!}{%
\begin{tabular}{@{}cllrrrrrrrrrr@{}}
\toprule[.15em]
\multicolumn{1}{c}{\textbf{\begin{tabular}[c]{@{}c@{}}Task\\Language\end{tabular}}} & \multicolumn{1}{c}{\textbf{\begin{tabular}[c]{@{}c@{}}Language\\ Representation\end{tabular}}} & \multicolumn{1}{c}{\textbf{\begin{tabular}[c]{@{}c@{}}Task\\ Representation\end{tabular}}} & \multicolumn{1}{c}{\textbf{bg}} & \multicolumn{1}{c}{\textbf{cs}} & \multicolumn{1}{c}{\textbf{el}} & \multicolumn{1}{c}{\textbf{ml}} & \multicolumn{1}{c}{\textbf{ro}} & \multicolumn{1}{c}{\textbf{sl}} & \multicolumn{1}{c}{\textbf{sk}} & \multicolumn{1}{c}{\textbf{sw}} & \multicolumn{1}{c}{\textbf{te}} & \multicolumn{1}{c}{\textbf{ur}} \\ \midrule[.15em]
\textbf{} & None & None & 0 & 0 & 0 & 0 & 0 & 0 & 0 & 0 & 0 & 0 \\ \midrule[.15em] 
\multirow{6}{*}{\textbf{ar}} & \multirow{2}{*}{None} & Adapter & 44.54 & 60.54 & 44.29 & 31.32 & 45.73 & {\ul 55.63} & 60.30 & 49.92 & {\ul 48.53} & {\ul 32.10} \\
 &  & Soft Prompt & 38.11 & 50.72 & 39.73 & 38.32 & 36.44 & 48.28 & 48.69 & 46.90 & 45.53 & 20.61 \\ \cmidrule(l){2-13}
 & \multirow{2}{*}{Adapter} & Adapter & {\ul 63.91} & \textbf{63.98} & \textbf{63.89} & \textbf{49.19} & 53.44 & 24.02 & {\ul 62.98} & 49.38 & 18.05 & 27.71 \\
 &  & Soft Prompt & \textbf{64.09} & 40.09 & 52.23 & {\ul 48.21} & \textbf{55.95} & 44.67 & 54.13 & {\ul 51.37} & 26.65 & 25.77 \\ \cmidrule(l){2-13}
 & \multirow{2}{*}{Soft Prompt} & Adapter & 53.66 & {\ul 62.72} & {\ul 52.68} & 41.95 & {\ul 54.82} & \textbf{60.70} & \textbf{64.62} & \textbf{55.43} & \textbf{51.43} & \textbf{43.36} \\
 &  & Soft Prompt & 48.84 & 40.30 & 46.02 & 42.93 & 33.25 & 37.20 & 44.48 & 30.41 & 46.01 & 25.18 \\ \midrule[.15em]
\multirow{6}{*}{\textbf{de}} & \multirow{2}{*}{None} & Adapter & 30.63 & {\ul 66.61} & 53.35 & 40.49 & \textbf{62.39} & 56.11 & 67.96 & {\ul 60.19} & 48.56 & 34.37 \\
 &  & Soft Prompt & 31.61 & 61.04 & 54.21 & {\ul 44.55} & 51.52 & 54.04 & 62.28 & 57.26 & 43.20 & 24.21 \\ \cmidrule(l){2-13}
 & \multirow{2}{*}{Adapter} & Adapter & \textbf{68.49} & 64.70 & \textbf{68.83} & \textbf{46.45} & 60.64 & 49.32 & 68.42 & \textbf{60.92} & 24.52 & 28.41 \\
 &  & Soft Prompt & {\ul 62.16} & 54.79 & 60.52 & 42.90 & 54.74 & 44.46 & 51.17 & 57.28 & 27.96 & \textbf{47.28} \\ \cmidrule(l){2-13}
 & \multirow{2}{*}{Soft Prompt} & Adapter & 39.84 & \textbf{67.12} & 55.29 & 40.43 & {\ul 61.59} & \textbf{63.28} & \textbf{71.04} & 56.53 & \textbf{49.70} & {\ul 45.00} \\
 &  & Soft Prompt & 43.30 & 62.84 & {\ul 61.20} & 44.44 & 52.13 & {\ul 57.21} & {\ul 70.80} & 54.24 & {\ul 48.67} & 34.37 \\ \midrule[.15em]
\multirow{6}{*}{\textbf{en}} & \multirow{2}{*}{None} & Adapter & 35.49 & 41.53 & {\ul 49.44} & 24.71 & 44.91 & 55.13 & 44.15 & 36.52 & 32.02 & 25.96 \\
 &  & Soft Prompt & 29.04 & {\ul 42.31} & 46.31 & {\ul 27.92} & {\ul 53.48} & 53.06 & 45.40 & 48.41 & {\ul 33.41} & {\ul 39.54} \\ \cmidrule(l){2-13} 
 & \multirow{2}{*}{Adapter} & Adapter & 21.42 & 33.99 & 39.86 & 17.16 & 42.90 & 23.39 & 23.18 & 32.18 & 13.91 & 21.95 \\
 &  & Soft Prompt & \textbf{50.16} & 40.62 & \textbf{56.07} & \textbf{37.99} & \textbf{63.94} & \textbf{56.46} & 35.97 & 45.38 & 19.98 & \textbf{61.68} \\ \cmidrule(l){2-13} 
 & \multirow{2}{*}{Soft Prompt} & Adapter & {\ul 44.12} & 40.65 & 47.99 & 24.97 & 52.36 & 51.04 & \textbf{51.62} & {\ul 48.55} & 31.23 & 23.32 \\
 &  & Soft Prompt & 22.90 & \textbf{44.33} & 41.92 & 25.22 & 49.67 & {\ul 55.77} & {\ul 48.64} & \textbf{48.77} & \textbf{33.85} & 9.87 \\ \midrule[.15em]
\multirow{6}{*}{\textbf{es}} & \multirow{2}{*}{None} & Adapter & 35.41 & \textbf{68.66} & 49.50 & 35.38 & \textbf{66.54} & {\ul 66.07} & \textbf{70.46} & {\ul 64.65} & 46.24 & 46.35 \\
 &  & Soft Prompt & 24.70 & 63.95 & 48.38 & 31.45 & 62.20 & 60.00 & 63.25 & 62.21 & {\ul 47.81} & 45.11 \\ \cmidrule(l){2-13}
 & \multirow{2}{*}{Adapter} & Adapter & {\ul 63.38} & 58.05 & \textbf{67.33} & {\ul 42.61} & 59.17 & 44.54 & 66.00 & 56.36 & 17.84 & 30.86 \\
 &  & Soft Prompt & \textbf{69.95} & 52.71 & 64.27 & \textbf{42.75} & 59.96 & 52.05 & 60.82 & 58.11 & 31.27 & {\ul 48.71} \\ \cmidrule(l){2-13}
 & \multirow{2}{*}{Soft Prompt} & Adapter & 54.37 & {\ul 67.45} & {\ul 64.81} & 40.37 & {\ul 64.62} & \textbf{70.83} & {\ul 68.78} & \textbf{66.53} & \textbf{50.91} & \textbf{53.78} \\
 &  & Soft Prompt & 44.72 & 65.55 & 39.99 & 39.17 & 63.38 & 63.52 & 66.04 & 61.86 & 40.59 & 47.45 \\ \midrule[.15em]
\multirow{6}{*}{\textbf{ru}} & \multirow{2}{*}{None} & Adapter & {\ul 72.90} & 46.64 & 34.14 & 25.54 & 49.66 & 45.84 & 48.16 & 50.02 & {\ul 49.73} & 37.42 \\
 &  & Soft Prompt & 69.54 & 53.60 & 25.83 & 34.23 & 51.11 & 51.04 & 51.10 & 48.37 & 49.36 & 40.56 \\ \cmidrule(l){2-13}
 & \multirow{2}{*}{Adapter} & Adapter & 71.70 & 54.14 & \textbf{72.40} & 38.30 & {\ul 59.91} & 30.15 & 50.78 & 48.08 & 18.41 & 25.92 \\
 &  & Soft Prompt & 73.07 & 53.33 & 65.15 & 37.04 & 53.23 & {\ul 58.78} & 52.96 & 50.20 & 21.37 & 27.01 \\ \cmidrule(l){2-13}
 & \multirow{2}{*}{Soft Prompt} & Adapter & \textbf{77.27} & {\ul 54.48} & 54.56 & {\ul 42.01} & \textbf{60.01} & 53.82 & {\ul 56.33} & \textbf{56.61} & \textbf{55.41} & \textbf{46.92} \\
 &  & Soft Prompt & 71.73 & \textbf{59.97} & {\ul 65.74} & \textbf{44.44} & 58.04 & \textbf{60.23} & \textbf{56.40} & {\ul 54.53} & 47.95 & {\ul 46.62} \\ \midrule[.15em]
\multirow{6}{*}{\textbf{zh}} & \multirow{2}{*}{None} & Adapter & 14.91 & 43.07 & 8.42 & 1.21 & 39.80 & 41.54 & 42.05 & 40.50 & 11.46 & 11.33 \\
 &  & Soft Prompt & 12.35 & {\ul 51.86} & 17.37 & 7.50 & {\ul 45.41} & \textbf{53.70} & {\ul 49.48} & 46.97 & 28.33 & 17.65 \\ \cmidrule(l){2-13}
 & \multirow{2}{*}{Adapter} & Adapter & {\ul 29.40} & 41.60 & \textbf{39.63} & 7.39 & 37.02 & 19.26 & 28.99 & 21.72 & 5.56 & 6.54 \\
 &  & Soft Prompt & \textbf{37.86} & 46.02 & 19.56 & \textbf{43.69} & 30.45 & 34.32 & 18.60 & \textbf{55.69} & \textbf{38.07} & \textbf{33.44} \\ \cmidrule(l){2-13}
 & \multirow{2}{*}{Soft Prompt} & Adapter & 28.41 & 46.21 & 24.62 & 8.49 & \textbf{50.18} & 44.69 & 47.55 & {\ul 51.13} & 28.76 & 15.06 \\
 &  & Soft Prompt & 23.34 & \textbf{55.43} & {\ul 32.02} & {\ul 12.41} & 43.96 & {\ul 53.55} & \textbf{55.26} & 48.51 & {\ul 32.11} & {\ul 18.78} \\ \bottomrule[.15em]
\end{tabular}}
\caption{Results for the named-entity recognition task using F1-Score. The best scores are \textbf{boldfaced}, and the second best are {\ul underlined}.}
\label{tab:ner-all}
\end{table*}

\begin{table*}[]
\centering
\small
\resizebox{\textwidth}{!}{%
\begin{tabular}{@{}cllrrrrrrrrrr@{}}
\toprule[.15em]
\multicolumn{1}{c}{\textbf{\begin{tabular}[c]{@{}c@{}}Task\\Language\end{tabular}}} & \multicolumn{1}{c}{\textbf{\begin{tabular}[c]{@{}c@{}}Language\\ Representation\end{tabular}}} & \multicolumn{1}{c}{\textbf{\begin{tabular}[c]{@{}c@{}}Task\\ Representation\end{tabular}}} & \multicolumn{1}{c}{\textbf{bg}} & \multicolumn{1}{c}{\textbf{cs}} & \multicolumn{1}{c}{\textbf{el}} & \multicolumn{1}{c}{\textbf{ml}} & \multicolumn{1}{c}{\textbf{ro}} & \multicolumn{1}{c}{\textbf{sl}} & \multicolumn{1}{c}{\textbf{sk}} & \multicolumn{1}{c}{\textbf{sw}} & \multicolumn{1}{c}{\textbf{te}} & \multicolumn{1}{c}{\textbf{ur}} \\ \midrule[.15em]
\textbf{} & None & None & 43.35 & 35.50 & 40.88 & 40.62 & 4.98 & 68.74 & 36.42 & 38.90 & 39.58 & 37.64 \\ \midrule[.15em]
 &  & Adapter & {\color[HTML]{009901} \textbf{74.77}} & {\color[HTML]{009901} 35.50} & {\color[HTML]{009901} {\ul 74.05}} & {\color[HTML]{009901} \textbf{68.98}} & {\color[HTML]{009901} 66.06} & {\color[HTML]{FE0000} 28.76} & {\color[HTML]{009901} 37.08} & {\color[HTML]{009901} \textbf{64.55}} & {\color[HTML]{009901} \textbf{66.67}} & {\color[HTML]{009901} \textbf{66.01}} \\
 & \multirow{-2}{*}{None} & Soft Prompt & {\color[HTML]{009901} 69.42} & {\color[HTML]{009901} 35.92} & {\color[HTML]{009901} 69.94} & {\color[HTML]{009901} 65.47} & {\color[HTML]{009901} \textbf{77.59}} & {\color[HTML]{FE0000} 25.45} & {\color[HTML]{FE0000} 35.83} & {\color[HTML]{009901} 59.94} & {\color[HTML]{009901} 64.23} & {\color[HTML]{009901} 60.68} \\ \cmidrule(l){2-13}
 &  & Adapter & {\color[HTML]{009901} 72.16} & {\color[HTML]{009901} 35.67} & {\color[HTML]{009901} \textbf{74.17}} & {\color[HTML]{009901} 65.59} & {\color[HTML]{009901} {\ul 77.15}} & {\color[HTML]{FE0000} 2.91} & {\color[HTML]{009901} {\ul 37.17}} & {\color[HTML]{009901} 62.18} & {\color[HTML]{009901} 55.93} & {\color[HTML]{009901} 47.17} \\
 & \multirow{-2}{*}{Adapter} & Soft Prompt & {\color[HTML]{009901} 56.39} & {\color[HTML]{009901} \textbf{38.08}} & {\color[HTML]{009901} 62.69} & {\color[HTML]{009901} 57.98} & {\color[HTML]{009901} 42.24} & {\color[HTML]{FE0000} \textbf{67.54}} & {\color[HTML]{009901} {\ul 37.17}} & {\color[HTML]{009901} 58.16} & {\color[HTML]{009901} 46.19} & {\color[HTML]{009901} 57.19} \\ \cmidrule(l){2-13}
 &  & Adapter & {\color[HTML]{009901} {\ul 74.21}} & {\color[HTML]{FE0000} 34.42} & {\color[HTML]{009901} 72.46} & {\color[HTML]{009901} {\ul 68.70}} & {\color[HTML]{009901} 63.62} & {\color[HTML]{FE0000} {\ul 32.77}} & {\color[HTML]{FE0000} 36.17} & {\color[HTML]{009901} {\ul 62.61}} & {\color[HTML]{009901} {\ul 66.05}} & {\color[HTML]{009901} {\ul 64.05}} \\
\multirow{-6}{*}{\textbf{ar}} & \multirow{-2}{*}{Soft Prompt} & Soft Prompt & {\color[HTML]{009901} 65.45} & {\color[HTML]{009901} {\ul 37.50}} & {\color[HTML]{009901} 67.84} & {\color[HTML]{009901} 62.53} & {\color[HTML]{009901} 59.03} & {\color[HTML]{FE0000} 30.96} & {\color[HTML]{009901} \textbf{37.25}} & {\color[HTML]{009901} 61.02} & {\color[HTML]{009901} 61.44} & {\color[HTML]{009901} 59.60} \\ \midrule[.15em]
 &  & Adapter & {\color[HTML]{009901} {\ul 75.03}} & {\color[HTML]{FE0000} 35.17} & {\color[HTML]{009901} \textbf{74.51}} & {\color[HTML]{009901} {\ul 69.92}} & {\color[HTML]{009901} 64.61} & {\color[HTML]{FE0000} 30.16} & {\color[HTML]{FE0000} 35.92} & {\color[HTML]{009901} \textbf{65.83}} & {\color[HTML]{009901} \textbf{68.90}} & {\color[HTML]{009901} \textbf{65.73}} \\
 & \multirow{-2}{*}{None} & Soft Prompt & {\color[HTML]{009901} 69.84} & {\color[HTML]{009901} 35.50} & {\color[HTML]{009901} 70.44} & {\color[HTML]{009901} 65.67} & {\color[HTML]{009901} \textbf{79.10}} & {\color[HTML]{FE0000} 31.36} & {\color[HTML]{009901} {\ul 37.00}} & {\color[HTML]{009901} 62.75} & {\color[HTML]{009901} 64.45} & {\color[HTML]{009901} 62.79} \\ \cmidrule(l){2-13}
 &  & Adapter & {\color[HTML]{009901} 74.41} & {\color[HTML]{FE0000} 34.92} & {\color[HTML]{009901} {\ul 72.34}} & {\color[HTML]{009901} 68.74} & {\color[HTML]{009901} 49.85} & {\color[HTML]{FE0000} 16.13} & {\color[HTML]{FE0000} 35.50} & {\color[HTML]{009901} 62.85} & {\color[HTML]{009901} 66.81} & {\color[HTML]{009901} 60.18} \\
 & \multirow{-2}{*}{Adapter} & Soft Prompt & {\color[HTML]{009901} 62.51} & {\color[HTML]{009901} \textbf{37.17}} & {\color[HTML]{009901} 62.81} & {\color[HTML]{009901} 55.29} & {\color[HTML]{009901} 35.16} & {\color[HTML]{FE0000} \textbf{46.19}} & {\color[HTML]{009901} 36.67} & {\color[HTML]{009901} 56.67} & {\color[HTML]{009901} 46.45} & {\color[HTML]{009901} 53.13} \\ \cmidrule(l){2-13}
 &  & Adapter & {\color[HTML]{009901} \textbf{75.23}} & {\color[HTML]{FE0000} 33.75} & {\color[HTML]{009901} 71.50} & {\color[HTML]{009901} \textbf{70.06}} & {\color[HTML]{009901} 60.25} & {\color[HTML]{FE0000} {\ul 35.17}} & {\color[HTML]{FE0000} 34.50} & {\color[HTML]{009901} {\ul 65.81}} & {\color[HTML]{009901} {\ul 67.13}} & {\color[HTML]{009901} {\ul 64.25}} \\
\multirow{-6}{*}{\textbf{de}} & \multirow{-2}{*}{Soft Prompt} & Soft Prompt & {\color[HTML]{009901} 69.12} & {\color[HTML]{009901} {\ul 36.67}} & {\color[HTML]{009901} 69.06} & {\color[HTML]{009901} 64.61} & {\color[HTML]{009901} {\ul 70.75}} & {\color[HTML]{FE0000} 26.35} & {\color[HTML]{009901} \textbf{37.67}} & {\color[HTML]{009901} 61.18} & {\color[HTML]{009901} 64.37} & {\color[HTML]{009901} 60.00} \\ \midrule[.15em]
 &  & Adapter & {\color[HTML]{009901} {\ul 75.03}} & {\color[HTML]{FE0000} 35.00} & {\color[HTML]{009901} \textbf{74.33}} & {\color[HTML]{009901} \textbf{70.30}} & {\color[HTML]{009901} 69.48} & {\color[HTML]{FE0000} 28.66} & {\color[HTML]{FE0000} 34.58} & {\color[HTML]{009901} \textbf{67.07}} & {\color[HTML]{009901} \textbf{69.38}} & {\color[HTML]{009901} \textbf{64.63}} \\
 & \multirow{-2}{*}{None} & Soft Prompt & {\color[HTML]{009901} 70.42} & {\color[HTML]{FE0000} 33.83} & {\color[HTML]{009901} 70.32} & {\color[HTML]{009901} 64.53} & {\color[HTML]{009901} 77.00} & {\color[HTML]{FE0000} 29.96} & {\color[HTML]{FE0000} 34.58} & {\color[HTML]{009901} 63.31} & {\color[HTML]{009901} 63.43} & {\color[HTML]{009901} 60.22} \\ \cmidrule(l){2-13}
 &  & Adapter & {\color[HTML]{009901} 64.07} & {\color[HTML]{FE0000} 35.33} & {\color[HTML]{009901} 64.79} & {\color[HTML]{009901} 56.51} & {\color[HTML]{009901} \textbf{90.09}} & {\color[HTML]{FE0000} 1.00} & {\color[HTML]{FE0000} 35.08} & {\color[HTML]{009901} 61.30} & {\color[HTML]{009901} 44.11} & {\color[HTML]{009901} 39.78} \\
 & \multirow{-2}{*}{Adapter} & Soft Prompt & {\color[HTML]{009901} 55.19} & {\color[HTML]{009901} \textbf{38.00}} & {\color[HTML]{009901} 51.30} & {\color[HTML]{009901} 46.53} & {\color[HTML]{009901} {\ul 81.35}} & {\color[HTML]{FE0000} 33.77} & {\color[HTML]{009901} \textbf{38.75}} & {\color[HTML]{009901} 56.97} & {\color[HTML]{FE0000} 33.33} & {\color[HTML]{009901} 52.81} \\ \cmidrule(l){2-13}
 &  & Adapter & {\color[HTML]{009901} \textbf{75.11}} & {\color[HTML]{FE0000} 33.83} & {\color[HTML]{009901} {\ul 72.50}} & {\color[HTML]{009901} {\ul 69.50}} & {\color[HTML]{009901} 54.05} & {\color[HTML]{FE0000} \textbf{35.07}} & {\color[HTML]{FE0000} 35.58} & {\color[HTML]{009901} {\ul 64.67}} & {\color[HTML]{009901} {\ul 67.88}} & {\color[HTML]{009901} {\ul 64.07}} \\
\multirow{-6}{*}{\textbf{en}} & \multirow{-2}{*}{Soft Prompt} & Soft Prompt & {\color[HTML]{009901} 65.37} & {\color[HTML]{009901} {\ul 35.58}} & {\color[HTML]{009901} 69.50} & {\color[HTML]{009901} 64.51} & {\color[HTML]{009901} 35.64} & {\color[HTML]{FE0000} {\ul 34.07}} & {\color[HTML]{009901} {\ul 37.08}} & {\color[HTML]{009901} 60.94} & {\color[HTML]{009901} 63.35} & {\color[HTML]{009901} 59.62} \\ \midrule[.15em]
 &  & Adapter & {\color[HTML]{009901} \textbf{74.87}} & {\color[HTML]{FE0000} 35.08} & {\color[HTML]{009901} \textbf{75.29}} & {\color[HTML]{009901} \textbf{70.40}} & {\color[HTML]{009901} 72.12} & {\color[HTML]{FE0000} 28.06} & {\color[HTML]{FE0000} 35.17} & {\color[HTML]{009901} \textbf{66.17}} & {\color[HTML]{009901} \textbf{70.00}} & {\color[HTML]{009901} \textbf{66.09}} \\
 & \multirow{-2}{*}{None} & Soft Prompt & {\color[HTML]{009901} 68.98} & {\color[HTML]{FE0000} 34.83} & {\color[HTML]{009901} 70.12} & {\color[HTML]{009901} 65.55} & {\color[HTML]{009901} {\ul 79.83}} & {\color[HTML]{FE0000} {\ul 31.76}} & {\color[HTML]{FE0000} 35.75} & {\color[HTML]{009901} 60.66} & {\color[HTML]{009901} 64.11} & {\color[HTML]{009901} 61.36} \\ \cmidrule(l){2-13}
 &  & Adapter & {\color[HTML]{009901} 73.95} & {\color[HTML]{FE0000} 34.92} & {\color[HTML]{009901} {\ul 74.61}} & {\color[HTML]{009901} 68.26} & {\color[HTML]{009901} 71.78} & {\color[HTML]{FE0000} 10.02} & {\color[HTML]{FE0000} 35.83} & {\color[HTML]{009901} {\ul 64.77}} & {\color[HTML]{009901} 59.40} & {\color[HTML]{009901} 50.42} \\
 & \multirow{-2}{*}{Adapter} & Soft Prompt & {\color[HTML]{009901} 54.97} & {\color[HTML]{009901} \textbf{38.75}} & {\color[HTML]{009901} 44.47} & {\color[HTML]{009901} 41.48} & {\color[HTML]{009901} \textbf{81.98}} & {\color[HTML]{FE0000} 8.42} & {\color[HTML]{009901} \textbf{36.42}} & {\color[HTML]{009901} 55.11} & {\color[HTML]{FE0000} 33.49} & {\color[HTML]{009901} 46.13} \\ \cmidrule(l){2-13}
 &  & Adapter & {\color[HTML]{009901} {\ul 74.27}} & {\color[HTML]{FE0000} 33.92} & {\color[HTML]{009901} 72.38} & {\color[HTML]{009901} {\ul 69.68}} & {\color[HTML]{009901} 67.58} & {\color[HTML]{FE0000} \textbf{32.67}} & {\color[HTML]{FE0000} 36.25} & {\color[HTML]{009901} 64.35} & {\color[HTML]{009901} {\ul 67.33}} & {\color[HTML]{009901} {\ul 62.95}} \\
\multirow{-6}{*}{\textbf{es}} & \multirow{-2}{*}{Soft Prompt} & Soft Prompt & {\color[HTML]{009901} 66.15} & {\color[HTML]{009901} {\ul 35.92}} & {\color[HTML]{009901} 67.54} & {\color[HTML]{009901} 62.77} & {\color[HTML]{009901} 78.52} & {\color[HTML]{FE0000} 28.76} & {\color[HTML]{FE0000} {\ul 36.33}} & {\color[HTML]{009901} 59.66} & {\color[HTML]{009901} 63.19} & {\color[HTML]{009901} 57.66} \\ \midrule[.15em]
 &  & Adapter & {\color[HTML]{009901} {\ul 75.55}} & {\color[HTML]{FE0000} 34.92} & {\color[HTML]{009901} \textbf{74.81}} & {\color[HTML]{009901} \textbf{70.56}} & {\color[HTML]{009901} 75.24} & {\color[HTML]{FE0000} \textbf{29.96}} & {\color[HTML]{FE0000} 34.67} & {\color[HTML]{009901} \textbf{67.54}} & {\color[HTML]{009901} \textbf{69.54}} & {\color[HTML]{009901} \textbf{66.29}} \\
 & \multirow{-2}{*}{None} & Soft Prompt & {\color[HTML]{009901} 69.90} & {\color[HTML]{009901} 36.25} & {\color[HTML]{009901} 69.78} & {\color[HTML]{009901} 64.85} & {\color[HTML]{009901} 82.62} & {\color[HTML]{FE0000} {\ul 28.16}} & {\color[HTML]{FE0000} 35.67} & {\color[HTML]{009901} 61.60} & {\color[HTML]{009901} 63.73} & {\color[HTML]{009901} 62.36} \\ \cmidrule(l){2-13}
 &  & Adapter & {\color[HTML]{009901} 72.63} & {\color[HTML]{009901} 35.75} & {\color[HTML]{009901} {\ul 74.19}} & {\color[HTML]{009901} 61.14} & {\color[HTML]{009901} {\ul 83.50}} & {\color[HTML]{FE0000} 4.91} & {\color[HTML]{FE0000} 36.25} & {\color[HTML]{009901} {\ul 65.77}} & {\color[HTML]{009901} 51.16} & {\color[HTML]{009901} 46.19} \\
 & \multirow{-2}{*}{Adapter} & Soft Prompt & {\color[HTML]{009901} 63.45} & {\color[HTML]{009901} \textbf{37.58}} & {\color[HTML]{009901} 53.43} & {\color[HTML]{009901} 44.51} & {\color[HTML]{009901} \textbf{87.01}} & {\color[HTML]{FE0000} 11.32} & {\color[HTML]{009901} \textbf{38.08}} & {\color[HTML]{009901} 54.67} & {\color[HTML]{FE0000} 37.60} & {\color[HTML]{009901} 53.13} \\ \cmidrule(l){2-13}
 &  & Adapter & {\color[HTML]{009901} \textbf{76.03}} & {\color[HTML]{FE0000} 34.92} & {\color[HTML]{009901} 73.75} & {\color[HTML]{009901} {\ul 68.18}} & {\color[HTML]{009901} 75.68} & {\color[HTML]{FE0000} 28.06} & {\color[HTML]{FE0000} 35.83} & {\color[HTML]{009901} {\ul 65.77}} & {\color[HTML]{009901} {\ul 68.24}} & {\color[HTML]{009901} {\ul 63.99}} \\
\multirow{-6}{*}{\textbf{ru}} & \multirow{-2}{*}{Soft Prompt} & Soft Prompt & {\color[HTML]{009901} 68.42} & {\color[HTML]{009901} {\ul 37.25}} & {\color[HTML]{009901} 67.72} & {\color[HTML]{009901} 63.53} & {\color[HTML]{009901} 75.05} & {\color[HTML]{FE0000} 23.65} & {\color[HTML]{009901} {\ul 37.25}} & {\color[HTML]{009901} 57.72} & {\color[HTML]{009901} 62.55} & {\color[HTML]{009901} 60.08} \\ \midrule[.15em]
 &  & Adapter & {\color[HTML]{009901} \textbf{74.95}} & {\color[HTML]{009901} \textbf{35.75}} & {\color[HTML]{009901} \textbf{73.79}} & {\color[HTML]{009901} \textbf{69.16}} & {\color[HTML]{009901} 74.32} & {\color[HTML]{FE0000} 28.66} & {\color[HTML]{FE0000} {\ul 36.33}} & {\color[HTML]{009901} \textbf{68.00}} & {\color[HTML]{009901} \textbf{69.38}} & {\color[HTML]{009901} \textbf{66.11}} \\
 & \multirow{-2}{*}{None} & Soft Prompt & {\color[HTML]{009901} 70.80} & {\color[HTML]{009901} {\ul 35.75}} & {\color[HTML]{009901} 70.44} & {\color[HTML]{009901} 66.59} & {\color[HTML]{009901} {\ul 78.52}} & {\color[HTML]{FE0000} 21.84} & {\color[HTML]{FE0000} 35.17} & {\color[HTML]{009901} 62.36} & {\color[HTML]{009901} 65.75} & {\color[HTML]{009901} 62.85} \\ \cmidrule(l){2-13}
 &  & Adapter & {\color[HTML]{009901} 61.72} & {\color[HTML]{FE0000} 34.08} & {\color[HTML]{009901} 64.27} & {\color[HTML]{009901} 50.90} & {\color[HTML]{009901} \textbf{89.01}} & {\color[HTML]{FE0000} 0.20} & {\color[HTML]{FE0000} 35.17} & {\color[HTML]{009901} 60.66} & {\color[HTML]{FE0000} 35.71} & {\color[HTML]{009901} 38.98} \\
 & \multirow{-2}{*}{Adapter} & Soft Prompt & {\color[HTML]{009901} 43.47} & {\color[HTML]{FE0000} 33.33} & {\color[HTML]{009901} 56.09} & {\color[HTML]{009901} 56.29} & {\color[HTML]{009901} 45.90} & {\color[HTML]{009901} \textbf{98.20}} & {\color[HTML]{FE0000} 35.00} & {\color[HTML]{009901} 47.07} & {\color[HTML]{FE0000} 34.57} & {\color[HTML]{009901} 47.76} \\ \cmidrule(l){2-13}
 &  & Adapter & {\color[HTML]{009901} {\ul 73.43}} & {\color[HTML]{FE0000} 35.42} & {\color[HTML]{009901} {\ul 72.16}} & {\color[HTML]{009901} {\ul 68.06}} & {\color[HTML]{009901} 65.77} & {\color[HTML]{FE0000} {\ul 31.06}} & {\color[HTML]{FE0000} 35.50} & {\color[HTML]{009901} {\ul 64.37}} & {\color[HTML]{009901} {\ul 66.91}} & {\color[HTML]{009901} {\ul 63.67}} \\
\multirow{-6}{*}{\textbf{zh}} & \multirow{-2}{*}{Soft Prompt} & Soft Prompt & {\color[HTML]{009901} 68.68} & {\color[HTML]{FE0000} 34.67} & {\color[HTML]{009901} 68.86} & {\color[HTML]{009901} 64.69} & {\color[HTML]{009901} 76.95} & {\color[HTML]{FE0000} 23.55} & {\color[HTML]{009901} \textbf{36.50}} & {\color[HTML]{009901} 60.42} & {\color[HTML]{009901} 63.39} & {\color[HTML]{009901} 61.78} \\  \bottomrule[.15em]
\end{tabular}}
 \caption{For NLI, we report accuracy as a metric. The best results for each language pair are highlighted in \textbf{bold} and the second best are {\ul underlined}. Additionally, language pairs with improved performance compared to inference-only are marked in {\color[HTML]{009901} green}, and those with decreased performance are marked in {\color[HTML]{FE0000} red}.}
\label{tab:nli-all}
\end{table*}

\begin{table*}[]
\centering
\small
\resizebox{\textwidth}{!}{%
\begin{tabular}{@{}cllrrrrrrrrrr@{}}
\toprule[.15em]
\multicolumn{1}{c}{\textbf{\begin{tabular}[c]{@{}c@{}}Task\\Language\end{tabular}}} & \multicolumn{1}{c}{\textbf{\begin{tabular}[c]{@{}c@{}}Language\\ Representation\end{tabular}}} & \multicolumn{1}{c}{\textbf{\begin{tabular}[c]{@{}c@{}}Task\\ Representation\end{tabular}}} & \multicolumn{1}{c}{\textbf{bg}} & \multicolumn{1}{c}{\textbf{cs}} & \multicolumn{1}{c}{\textbf{el}} & \multicolumn{1}{c}{\textbf{ml}} & \multicolumn{1}{c}{\textbf{ro}} & \multicolumn{1}{c}{\textbf{sl}} & \multicolumn{1}{c}{\textbf{sk}} & \multicolumn{1}{c}{\textbf{sw}} & \multicolumn{1}{c}{\textbf{te}} & \multicolumn{1}{c}{\textbf{ur}} \\ \midrule[.15em]
\textbf{} & None & None & 0 & 0 & 0 & 0 & 0 & 0 & 0 & 0 & 0 & 0 \\ \midrule[.15em]
\multirow{6}{*}{\textbf{ar}} & \multirow{2}{*}{None} & Adapter & 84.78 & \textbf{70.74} & {\ul 86.77} & {\ul 86.45} & 86.06 & 82.74 & \textbf{72.52} & \textbf{86.59} & {\ul 84.32} & 81.50 \\
 &  & Soft Prompt & {\ul 86.48} & 63.08 & 84.90 & 86.32 & {\ul 87.61} & 80.77 & 56.91 & 81.47 & \textbf{84.47} & \textbf{81.92} \\ \cmidrule(l){2-13}
 & \multirow{2}{*}{Adapter} & Adapter & 58.38 & 45.01 & 74.35 & 83.39 & 73.17 & 71.29 & 36.69 & 82.63 & 77.91 & 80.18 \\
 &  & Soft Prompt & \textbf{89.22} & 56.49 & 85.76 & \textbf{86.82} & 78.56 & 82.24 & 48.71 & {\ul 85.96} & 82.44 & 81.62 \\ \cmidrule(l){2-13}
 & \multirow{2}{*}{Soft Prompt} & Adapter & 85.83 & {\ul 67.14} & \textbf{87.29} & 82.13 & \textbf{88.41} & \textbf{85.08} & {\ul 68.66} & 84.36 & 82.47 & 78.36 \\
 &  & Soft Prompt & 85.89 & 61.41 & 81.91 & 85.13 & 86.84 & {\ul 84.60} & 57.40 & 84.20 & 81.95 & {\ul 81.71} \\ \midrule[.15em]
\multirow{6}{*}{\textbf{de}} & \multirow{2}{*}{None} & Adapter & \textbf{97.63} & 69.72 & \textbf{97.74} & 94.66 & \textbf{98.11} & \textbf{97.11} & 73.70 & \textbf{97.25} & \textbf{93.73} & \textbf{93.17} \\
 &  & Soft Prompt & 94.49 & 65.32 & 96.42 & \textbf{96.28} & 92.33 & 88.16 & 62.31 & 90.69 & 91.54 & {\ul 89.10} \\ \cmidrule(l){2-13}
 & \multirow{2}{*}{Adapter} & Adapter & 97.03 & \textbf{78.84} & 95.30 & 85.30 & 97.19 & 95.28 & {\ul 82.91} & 90.39 & 79.96 & 85.29 \\
 &  & Soft Prompt & 93.15 & 58.68 & 94.46 & {\ul 95.17} & 95.57 & 96.11 & 67.99 & 87.70 & {\ul 93.06} & 65.98 \\ \cmidrule(l){2-13}
 & \multirow{2}{*}{Soft Prompt} & Adapter & {\ul 97.06} & {\ul 74.94} & {\ul 96.84} & 92.17 & {\ul 97.76} & {\ul 96.85} & \textbf{82.96} & {\ul 95.40} & 89.64 & 86.71 \\
 &  & Soft Prompt & 94.51 & 63.87 & 91.18 & 94.29 & 95.44 & 94.49 & 57.84 & 86.67 & 93.02 & 76.01 \\ \midrule[.15em]
\multirow{6}{*}{\textbf{en}} & \multirow{2}{*}{None} & Adapter & {\ul 95.79} & \textbf{85.10} & {\ul 97.46} & 89.15 & 94.52 & 89.31 & 83.09 & \textbf{91.98} & {\ul 87.46} & {\ul 88.65} \\
 &  & Soft Prompt & 94.98 & 55.29 & 95.09 & \textbf{93.12} & 91.50 & 91.47 & 53.03 & 91.17 & \textbf{93.50} & 88.35 \\ \cmidrule(l){2-13}
 & \multirow{2}{*}{Adapter} & Adapter & \textbf{97.22} & 82.95 & \textbf{98.04} & {\ul 92.28} & \textbf{96.98} & \textbf{95.80} & \textbf{85.98} & {\ul 91.20} & 85.99 & \textbf{89.57} \\
 &  & Soft Prompt & 88.69 & 40.60 & 94.31 & 86.24 & {\ul 94.97} & {\ul 93.09} & 46.60 & 88.02 & 81.91 & 86.00 \\ \cmidrule(l){2-13}
 & \multirow{2}{*}{Soft Prompt} & Adapter & 87.61 & {\ul 83.47} & 95.89 & 77.89 & 88.38 & 81.98 & {\ul 85.49} & 87.42 & 77.34 & 81.67 \\
 &  & Soft Prompt & 80.71 & 75.72 & 92.71 & 82.49 & 90.44 & 84.21 & 70.08 & 82.65 & 79.55 & 75.16 \\ \midrule[.15em]
\multirow{6}{*}{\textbf{es}} & \multirow{2}{*}{None} & Adapter & 95.66 & 76.98 & 95.85 & 91.93 & \textbf{97.06} & {\ul 95.81} & 74.40 & \textbf{95.62} & {\ul 93.15} & \textbf{90.57} \\
 &  & Soft Prompt & 92.80 & 76.38 & 94.59 & 91.93 & 90.62 & 82.23 & 74.78 & 92.40 & 91.90 & 83.25 \\ \cmidrule(l){2-13}
 & \multirow{2}{*}{Adapter} & Adapter & 84.35 & 73.15 & {\ul 95.94} & 92.67 & 96.16 & 94.07 & {\ul 81.65} & 93.48 & \textbf{93.80} & {\ul 88.13} \\
 &  & Soft Prompt & 92.98 & 77.88 & 96.15 & {\ul 93.27} & 96.03 & 93.07 & 81.02 & 92.09 & 83.19 & 71.66 \\ \cmidrule(l){2-13}
 & \multirow{2}{*}{Soft Prompt} & Adapter & \textbf{97.47} & \textbf{87.14} & \textbf{97.41} & \textbf{94.78} & {\ul 96.89} & \textbf{96.01} & \textbf{86.36} & {\ul 95.55} & 91.72 & 86.85 \\
 &  & Soft Prompt & {\ul 96.35} & {\ul 79.97} & 95.79 & 90.56 & 95.08 & 92.59 & 78.85 & 92.85 & 89.11 & 87.56 \\ \midrule[.15em]
\multirow{6}{*}{\textbf{ru}} & \multirow{2}{*}{None} & Adapter & {\ul 98.87} & {\ul 72.15} & \textbf{98.46} & 90.34 & {\ul 98.26} & 96.35 & {\ul 74.45} & \textbf{97.10} & 89.87 & \textbf{92.93} \\
 &  & Soft Prompt & 98.33 & 58.26 & 96.09 & {\ul 95.24} & 87.02 & 84.39 & 59.06 & 92.15 & \textbf{95.03} & 92.11 \\ \cmidrule(l){2-13}
 & \multirow{2}{*}{Adapter} & Adapter & 97.90 & \textbf{88.93} & 94.70 & 82.33 & 96.89 & 92.79 & \textbf{88.80} & 91.75 & 77.95 & 87.34 \\
 &  & Soft Prompt & 97.36 & 13.87 & 96.77 & \textbf{95.50} & 95.03 & 95.70 & 20.58 & 95.59 & 91.23 & {\ul 92.34} \\ \cmidrule(l){2-13}
 & \multirow{2}{*}{Soft Prompt} & Adapter & 98.59 & 67.29 & {\ul 97.46} & 88.39 & \textbf{98.53} & \textbf{97.73} & 74.44 & {\ul 95.83} & 88.39 & 91.36 \\
 &  & Soft Prompt & \textbf{98.89} & 48.65 & 96.71 & 93.57 & 97.32 & {\ul 96.85} & 38.89 & 95.17 & {\ul 92.26} & 89.83 \\ \midrule[.15em]
\multirow{6}{*}{\textbf{zh}} & \multirow{2}{*}{None} & Adapter & 93.70 & 74.08 & {\ul 95.07} & 90.97 & 93.44 & 85.53 & 74.05 & {\ul 90.99} & 88.47 & {\ul 90.22} \\
 &  & Soft Prompt & 91.44 & 66.83 & 92.58 & \textbf{94.18} & 84.92 & 77.69 & 66.55 & 74.85 & 90.33 & 81.90 \\ \cmidrule(l){2-13}
 & \multirow{2}{*}{Adapter} & Adapter & \textbf{97.08} & \textbf{84.02} & \textbf{97.09} & 90.20 & \textbf{97.27} & \textbf{95.58} & \textbf{84.14} & 90.62 & 90.89 & \textbf{90.58} \\
 &  & Soft Prompt & 66.88 & 69.81 & 65.43 & 67.65 & 67.11 & 66.28 & 67.57 & 65.58 & 65.92 & 65.92 \\ \cmidrule(l){2-13}
 & \multirow{2}{*}{Soft Prompt} & Adapter & {\ul 96.07} & {\ul 76.96} & 93.18 & 89.41 & {\ul 95.15} & {\ul 94.37} & {\ul 80.19} & \textbf{91.13} & {\ul 91.22} & 82.55 \\
 &  & Soft Prompt & 95.30 & 73.52 & 89.10 & {\ul 94.01} & 92.61 & 90.44 & 62.60 & 83.33 & \textbf{91.33} & 85.38 \\\bottomrule[.15em]
\end{tabular}}
\caption{Results for the check-worthy claim detection task for cross-lingual transfer. Results are reported using F1-Score, with the best scores in \textbf{bold} and the second best {\ul underlined}.}
\label{tab:cwcd-all}
\end{table*}

\section{Evaluation with Multiple Training Seeds}
\label{app:multiple-seeds}

In Table~\ref{tab:seeds}, we report the evaluation results of all configurations that were trained on the German version of the WikiANN dataset using three different seeds. Along with the mean values, we also report the standard deviation

The obtained results demonstrate that the best results for knowledge transfer from German to other languages are obtained by using task adapters for Bulgarian, Greek, Malayalam, Romanian and Swahili. In contrast, the best combination for Czech, Slovenian, Slovak, Telugu and Urdu was a soft language prompt with a task adapter. This observation supports our previous findings that both configurations achieved superior results on the NER task when transferring knowledge from German.

\begin{table*}[]
\centering
\small
\resizebox{\textwidth}{!}{%
\begin{tabular}{@{}llllllllllll@{}}
\toprule[.15em]
\multicolumn{1}{c}{\textbf{\begin{tabular}[c]{@{}c@{}}Language\\ Representation\end{tabular}}} & \multicolumn{1}{c}{\textbf{\begin{tabular}[c]{@{}c@{}}Task \\ Representation\end{tabular}}} & \multicolumn{1}{c}{\textbf{bg}} & \multicolumn{1}{c}{\textbf{cs}} & \multicolumn{1}{c}{\textbf{el}} & \multicolumn{1}{c}{\textbf{ml}} & \multicolumn{1}{c}{\textbf{ro}} & \multicolumn{1}{c}{\textbf{sl}} & \multicolumn{1}{c}{\textbf{sk}} & \multicolumn{1}{c}{\textbf{sw}} & \multicolumn{1}{c}{\textbf{te}} & \multicolumn{1}{c}{\textbf{ur}} \\ \midrule[.15em]
\multirow{2}{*}{None} & Adapter & 38.46 ± 9.60 &   {\ul 66.00} ± 0.77 &  55.88 ± 5.33 &  40.99 ± 0.67 &  60.97 ± 1.74 &  56.29 ± 0.62 &  67.55 ± 1.22 &  {\ul 58.60} ± 1.97 &  47.69 ± 2.35 &   29.15 ± 8.78 \\ \cmidrule(l){2-12} 
 & Soft Prompt & 28.98 ± 3.24 &   61.48 ± 0.69 &  50.64 ± 6.63 &  43.08 ± 1.81 &  52.65 ± 1.58 &  54.03 ± 0.98 &  62.42 ± 0.75 &  56.49 ± 2.86 &  43.94 ± 1.64 &   26.67 ± 3.05 \\ \midrule[.15em]
\multirow{2}{*}{Adapter} & Adapter & \textbf{66.85} ± 2.03 &   64.78 ± 0.15 &  \textbf{69.13} ± 1.48 &  \textbf{47.34} ± 1.86 &  \textbf{64.07} ± 4.97 &  49.29 ± 2.42 &  68.15 ± 1.19 &  \textbf{59.45} ± 2.28 &  25.96 ± 2.86 &  20.20 ± 11.17 \\ \cmidrule(l){2-12} 
 & Soft Prompt & {\ul 63.93} ± 3.01 &  50.76 ± 14.17 &  62.05 ± 3.25 &  43.61 ± 2.42 &  56.11 ± 3.25 &  50.30 ± 7.88 &  53.63 ± 4.44 &  55.27 ± 3.95 &  26.83 ± 4.04 &   {\ul 46.63} ± 1.51 \\ \midrule[.15em]
\multirow{2}{*}{Soft Prompt} & Adapter & 42.19 ± 2.93 &   \textbf{67.25} ± 0.23 &  60.19 ± 6.02 &  43.67 ± 4.32 &  {\ul 61.89} ± 1.73 &  \textbf{61.75} ± 1.90 &  \textbf{72.95} ± 2.35 &  56.29 ± 0.65 &  \textbf{52.58} ± 5.43 &   \textbf{47.86} ± 3.77 \\ \cmidrule(l){2-12} 
 & Soft Prompt & 50.66 ± 12.65 &   64.00 ± 2.18 &  {\ul 63.38} ± 3.85 &  {\ul 46.37} ± 2.46 &  55.90 ± 4.95 &  {\ul 57.77} ± 0.71 &  {\ul 69.87} ± 2.21 &  54.26 ± 1.31 &  {\ul 48.35} ± 2.19 &   37.46 ± 3.93 \\ \bottomrule[.15em]
\end{tabular}}
\caption{Results of cross-lingual transfer from German to six languages for the NER task. We report the mean of three runs along with the standard deviation. The best results are \textbf{bolded} and the second best results are {\ul underlined}.}
\label{tab:seeds}
\end{table*}

%% file: acl_latex.bbl
\begin{thebibliography}{59}
\providecommand{\natexlab}[1]{#1}

\bibitem[{cs-()}]{cs-anli}

\newblock {CS ANLI}.
\newblock \url{https://huggingface.co/datasets/ctu-aic/anli_cs}.
\newblock Accessed: 2024-05-30.

\bibitem[{Aggarwal et~al.(2022)Aggarwal, Gupta, and Kunchukuttan}]{aggarwal-etal-2022-indicxnli}
Divyanshu Aggarwal, Vivek Gupta, and Anoop Kunchukuttan. 2022.
\newblock \href {https://doi.org/10.18653/v1/2022.emnlp-main.755} {{I}ndic{XNLI}: Evaluating multilingual inference for {I}ndian languages}.
\newblock In \emph{Proceedings of the 2022 Conference on Empirical Methods in Natural Language Processing}, pages 10994--11006, Abu Dhabi, United Arab Emirates. Association for Computational Linguistics.

\bibitem[{Alabi et~al.(2024)Alabi, Mosbach, Eyal, Klakow, and Geva}]{alabi-etal-2024-hidden}
Jesujoba Alabi, Marius Mosbach, Matan Eyal, Dietrich Klakow, and Mor Geva. 2024.
\newblock \href {https://doi.org/10.18653/v1/2024.acl-long.356} {The hidden space of transformer language adapters}.
\newblock In \emph{Proceedings of the 62nd Annual Meeting of the Association for Computational Linguistics (Volume 1: Long Papers)}, pages 6588--6607, Bangkok, Thailand. Association for Computational Linguistics.

\bibitem[{Ansell et~al.(2022)Ansell, Ponti, Korhonen, and Vuli{\'c}}]{ansell-etal-2022-composable}
Alan Ansell, Edoardo Ponti, Anna Korhonen, and Ivan Vuli{\'c}. 2022.
\newblock \href {https://doi.org/10.18653/v1/2022.acl-long.125} {Composable sparse fine-tuning for cross-lingual transfer}.
\newblock In \emph{Proceedings of the 60th Annual Meeting of the Association for Computational Linguistics (Volume 1: Long Papers)}, pages 1778--1796, Dublin, Ireland. Association for Computational Linguistics.

\bibitem[{Ansell et~al.(2021)Ansell, Ponti, Pfeiffer, Ruder, Glava{\v{s}}, Vuli{\'c}, and Korhonen}]{ansell-etal-2021-mad-g}
Alan Ansell, Edoardo~Maria Ponti, Jonas Pfeiffer, Sebastian Ruder, Goran Glava{\v{s}}, Ivan Vuli{\'c}, and Anna Korhonen. 2021.
\newblock \href {https://doi.org/10.18653/v1/2021.findings-emnlp.410} {{MAD}-{G}: {M}ultilingual adapter generation for efficient cross-lingual transfer}.
\newblock In \emph{Findings of the Association for Computational Linguistics: EMNLP 2021}, pages 4762--4781, Punta Cana, Dominican Republic. Association for Computational Linguistics.

\bibitem[{Arif et~al.(2024)Arif, Farid, Athar, and Raza}]{arif-etal-2024-uqa}
Samee Arif, Sualeha Farid, Awais Athar, and Agha~Ali Raza. 2024.
\newblock \href {https://aclanthology.org/2024.lrec-main.1497} {{UQA}: Corpus for {U}rdu question answering}.
\newblock In \emph{Proceedings of the 2024 Joint International Conference on Computational Linguistics, Language Resources and Evaluation (LREC-COLING 2024)}, pages 17237--17244, Torino, Italia. ELRA and ICCL.

\bibitem[{Artetxe et~al.(2020)Artetxe, Ruder, and Yogatama}]{artetxe-etal-2020-cross}
Mikel Artetxe, Sebastian Ruder, and Dani Yogatama. 2020.
\newblock \href {https://doi.org/10.18653/v1/2020.acl-main.421} {On the cross-lingual transferability of monolingual representations}.
\newblock In \emph{Proceedings of the 58th Annual Meeting of the Association for Computational Linguistics}, pages 4623--4637, Online. Association for Computational Linguistics.

\bibitem[{Asai et~al.(2022)Asai, Salehi, Peters, and Hajishirzi}]{asai-etal-2022-attempt}
Akari Asai, Mohammadreza Salehi, Matthew Peters, and Hannaneh Hajishirzi. 2022.
\newblock \href {https://doi.org/10.18653/v1/2022.emnlp-main.446} {{ATTEMPT}: Parameter-efficient multi-task tuning via attentional mixtures of soft prompts}.
\newblock In \emph{Proceedings of the 2022 Conference on Empirical Methods in Natural Language Processing}, pages 6655--6672, Abu Dhabi, United Arab Emirates. Association for Computational Linguistics.

\bibitem[{Borovi{\v{c}} et~al.(2022)Borovi{\v{c}}, {\v{Z}}agar, Ferme, Majninger, Ojster{\v{s}}ek, {\v{S}}majdek, Zirkelbach, Zupani{\v{c}}, Jazbin{\v{s}}ek, {\v{Z}}itnik et~al.}]{borovivc2022slovene}
Mladen Borovi{\v{c}}, Kristjan {\v{Z}}agar, Marko Ferme, Sandi Majninger, Milan Ojster{\v{s}}ek, Uro{\v{s}} {\v{S}}majdek, Maj Zirkelbach, Matja{\v{z}} Zupani{\v{c}}, Meta Jazbin{\v{s}}ek, Slavko {\v{Z}}itnik, et~al. 2022.
\newblock Slovene translation of the squad2. 0 dataset.

\bibitem[{Conneau et~al.(2020)Conneau, Khandelwal, Goyal, Chaudhary, Wenzek, Guzm{\'a}n, Grave, Ott, Zettlemoyer, and Stoyanov}]{conneau-etal-2020-unsupervised}
Alexis Conneau, Kartikay Khandelwal, Naman Goyal, Vishrav Chaudhary, Guillaume Wenzek, Francisco Guzm{\'a}n, Edouard Grave, Myle Ott, Luke Zettlemoyer, and Veselin Stoyanov. 2020.
\newblock \href {https://doi.org/10.18653/v1/2020.acl-main.747} {Unsupervised cross-lingual representation learning at scale}.
\newblock In \emph{Proceedings of the 58th Annual Meeting of the Association for Computational Linguistics}, pages 8440--8451, Online. Association for Computational Linguistics.

\bibitem[{Conneau et~al.(2018)Conneau, Rinott, Lample, Williams, Bowman, Schwenk, and Stoyanov}]{conneau2018xnli}
Alexis Conneau, Ruty Rinott, Guillaume Lample, Adina Williams, Samuel~R. Bowman, Holger Schwenk, and Veselin Stoyanov. 2018.
\newblock Xnli: Evaluating cross-lingual sentence representations.
\newblock In \emph{Proceedings of the 2018 Conference on Empirical Methods in Natural Language Processing}. Association for Computational Linguistics.

\bibitem[{Dettmers et~al.(2023)Dettmers, Pagnoni, Holtzman, and Zettlemoyer}]{NEURIPS2023_1feb8787}
Tim Dettmers, Artidoro Pagnoni, Ari Holtzman, and Luke Zettlemoyer. 2023.
\newblock \href {https://proceedings.neurips.cc/paper_files/paper/2023/file/1feb87871436031bdc0f2beaa62a049b-Paper-Conference.pdf} {Qlora: Efficient finetuning of quantized llms}.
\newblock In \emph{Advances in Neural Information Processing Systems}, volume~36, pages 10088--10115. Curran Associates, Inc.

\bibitem[{Doddapaneni et~al.(2023)Doddapaneni, Aralikatte, Ramesh, Goyal, Khapra, Kunchukuttan, and Kumar}]{doddapaneni-etal-2023-towards}
Sumanth Doddapaneni, Rahul Aralikatte, Gowtham Ramesh, Shreya Goyal, Mitesh~M. Khapra, Anoop Kunchukuttan, and Pratyush Kumar. 2023.
\newblock \href {https://doi.org/10.18653/v1/2023.acl-long.693} {Towards leaving no {I}ndic language behind: Building monolingual corpora, benchmark and models for {I}ndic languages}.
\newblock In \emph{Proceedings of the 61st Annual Meeting of the Association for Computational Linguistics (Volume 1: Long Papers)}, pages 12402--12426, Toronto, Canada. Association for Computational Linguistics.

\bibitem[{He et~al.(2022)He, Zhou, Ma, Berg-Kirkpatrick, and Neubig}]{he2022unifiedviewparameterefficienttransfer}
Junxian He, Chunting Zhou, Xuezhe Ma, Taylor Berg-Kirkpatrick, and Graham Neubig. 2022.
\newblock \href {https://arxiv.org/abs/2110.04366} {Towards a unified view of parameter-efficient transfer learning}.
\newblock \emph{Preprint}, arXiv:2110.04366.

\bibitem[{Hl{\'a}dek et~al.(2023)Hl{\'a}dek, Sta{\v{s}}, Juh{\'a}r, and Koct{\'u}r}]{hladek2023slovak}
Daniel Hl{\'a}dek, J{\'a}n Sta{\v{s}}, Jozef Juh{\'a}r, and Tom{\'a}{\v{s}} Koct{\'u}r. 2023.
\newblock Slovak dataset for multilingual question answering.
\newblock \emph{IEEE Access}, 11:32869--32881.

\bibitem[{Houlsby et~al.(2019{\natexlab{a}})Houlsby, Giurgiu, Jastrzebski, Morrone, De~Laroussilhe, Gesmundo, Attariyan, and Gelly}]{houlsby2019parameter}
Neil Houlsby, Andrei Giurgiu, Stanislaw Jastrzebski, Bruna Morrone, Quentin De~Laroussilhe, Andrea Gesmundo, Mona Attariyan, and Sylvain Gelly. 2019{\natexlab{a}}.
\newblock Parameter-efficient transfer learning for nlp.
\newblock In \emph{International conference on machine learning}, pages 2790--2799. PMLR.

\bibitem[{Houlsby et~al.(2019{\natexlab{b}})Houlsby, Giurgiu, Jastrzebski, Morrone, De~Laroussilhe, Gesmundo, Attariyan, and Gelly}]{pmlr-v97-houlsby19a}
Neil Houlsby, Andrei Giurgiu, Stanislaw Jastrzebski, Bruna Morrone, Quentin De~Laroussilhe, Andrea Gesmundo, Mona Attariyan, and Sylvain Gelly. 2019{\natexlab{b}}.
\newblock \href {https://proceedings.mlr.press/v97/houlsby19a.html} {Parameter-efficient transfer learning for {NLP}}.
\newblock In \emph{Proceedings of the 36th International Conference on Machine Learning}, volume~97 of \emph{Proceedings of Machine Learning Research}, pages 2790--2799. PMLR.

\bibitem[{Hu et~al.(2020)Hu, Ruder, Siddhant, Neubig, Firat, and Johnson}]{pmlr-v119-hu20b}
Junjie Hu, Sebastian Ruder, Aditya Siddhant, Graham Neubig, Orhan Firat, and Melvin Johnson. 2020.
\newblock \href {https://proceedings.mlr.press/v119/hu20b.html} {{XTREME}: A massively multilingual multi-task benchmark for evaluating cross-lingual generalisation}.
\newblock In \emph{Proceedings of the 37th International Conference on Machine Learning}, volume 119 of \emph{Proceedings of Machine Learning Research}, pages 4411--4421. PMLR.

\bibitem[{Huang et~al.(2022)Huang, Ma, Zhang, Wei, and Wang}]{huang-etal-2022-zero}
Lianzhe Huang, Shuming Ma, Dongdong Zhang, Furu Wei, and Houfeng Wang. 2022.
\newblock \href {https://doi.org/10.18653/v1/2022.emnlp-main.790} {Zero-shot cross-lingual transfer of prompt-based tuning with a unified multilingual prompt}.
\newblock In \emph{Proceedings of the 2022 Conference on Empirical Methods in Natural Language Processing}, pages 11488--11497, Abu Dhabi, United Arab Emirates. Association for Computational Linguistics.

\bibitem[{Hyben et~al.(2023)Hyben, Kula, Srba, Moro, and Simko}]{hyben2023bigger}
Martin Hyben, Sebastian Kula, Ivan Srba, Robert Moro, and Jakub Simko. 2023.
\newblock \href {https://arxiv.org/abs/2311.06121} {Is it indeed bigger better? the comprehensive study of claim detection lms applied for disinformation tackling}.
\newblock \emph{Preprint}, arXiv:2311.06121.

\bibitem[{Klemen et~al.(2024)Klemen, {\v{Z}}agar, {\v{C}}ibej, and Robnik-{\v{S}}ikonja}]{klemen-etal-2024-si}
Matej Klemen, Ale{\v{s}} {\v{Z}}agar, Jaka {\v{C}}ibej, and Marko Robnik-{\v{S}}ikonja. 2024.
\newblock \href {https://aclanthology.org/2024.lrec-main.1294} {{SI}-{NLI}: A {S}lovene natural language inference dataset and its evaluation}.
\newblock In \emph{Proceedings of the 2024 Joint International Conference on Computational Linguistics, Language Resources and Evaluation (LREC-COLING 2024)}, pages 14859--14870, Torino, Italia. ELRA and ICCL.

\bibitem[{Klimaszewski et~al.(2024)Klimaszewski, Andruszkiewicz, and Birch}]{klimaszewski2024train}
Mateusz Klimaszewski, Piotr Andruszkiewicz, and Alexandra Birch. 2024.
\newblock \href {https://arxiv.org/abs/2404.15737} {No train but gain: Language arithmetic for training-free language adapters enhancement}.
\newblock \emph{Preprint}, arXiv:2404.15737.

\bibitem[{Kunz and Holmström(2024)}]{kunz2024impact}
Jenny Kunz and Oskar Holmström. 2024.
\newblock \href {https://arxiv.org/abs/2402.00149} {The impact of language adapters in cross-lingual transfer for nlu}.
\newblock \emph{Preprint}, arXiv:2402.00149.

\bibitem[{Lee et~al.(2022)Lee, Hwang, and Kim}]{lee-etal-2022-fad}
Jaeseong Lee, Seung-won Hwang, and Taesup Kim. 2022.
\newblock \href {https://aclanthology.org/2022.aacl-short.8} {{FAD}-{X}: Fusing adapters for cross-lingual transfer to low-resource languages}.
\newblock In \emph{Proceedings of the 2nd Conference of the Asia-Pacific Chapter of the Association for Computational Linguistics and the 12th International Joint Conference on Natural Language Processing (Volume 2: Short Papers)}, pages 57--64, Online only. Association for Computational Linguistics.

\bibitem[{Lester et~al.(2021)Lester, Al-Rfou, and Constant}]{lester-etal-2021-power}
Brian Lester, Rami Al-Rfou, and Noah Constant. 2021.
\newblock \href {https://doi.org/10.18653/v1/2021.emnlp-main.243} {The power of scale for parameter-efficient prompt tuning}.
\newblock In \emph{Proceedings of the 2021 Conference on Empirical Methods in Natural Language Processing}, pages 3045--3059, Online and Punta Cana, Dominican Republic. Association for Computational Linguistics.

\bibitem[{Lewis et~al.(2019)Lewis, Oguz, Rinott, Riedel, and Schwenk}]{lewis2019mlqa}
Patrick Lewis, Barlas Oguz, Ruty Rinott, Sebastian Riedel, and Holger Schwenk. 2019.
\newblock Mlqa: Evaluating cross-lingual extractive question answering.
\newblock \emph{arXiv preprint arXiv:1910.07475}, arXiv: 1910.07475.

\bibitem[{Liu and Huang(2023)}]{10.1145/3539618.3592043}
Lei Liu and Jimmy~Xiangji Huang. 2023.
\newblock \href {https://doi.org/10.1145/3539618.3592043} {Prompt learning to mitigate catastrophic forgetting in cross-lingual transfer for open-domain dialogue generation}.
\newblock In \emph{Proceedings of the 46th International ACM SIGIR Conference on Research and Development in Information Retrieval}, SIGIR '23, page 2287–2292, New York, NY, USA. Association for Computing Machinery.

\bibitem[{Luo et~al.(2024)Luo, Yang, Meng, Li, Zhou, and Zhang}]{luo2024empirical}
Yun Luo, Zhen Yang, Fandong Meng, Yafu Li, Jie Zhou, and Yue Zhang. 2024.
\newblock \href {https://arxiv.org/abs/2308.08747} {An empirical study of catastrophic forgetting in large language models during continual fine-tuning}.
\newblock \emph{Preprint}, arXiv:2308.08747.

\bibitem[{Mackov{\'a} and Straka(2020)}]{10.1007/978-3-030-58323-1_18}
Kate{\v{r}}ina Mackov{\'a} and Milan Straka. 2020.
\newblock Reading comprehension in czech via machine translation and cross-lingual transfer.
\newblock In \emph{Text, Speech, and Dialogue}, pages 171--179, Cham. Springer International Publishing.

\bibitem[{McCloskey and Cohen(1989)}]{MCCLOSKEY1989109}
Michael McCloskey and Neal~J. Cohen. 1989.
\newblock \href {https://doi.org/10.1016/S0079-7421(08)60536-8} {Catastrophic interference in connectionist networks: The sequential learning problem}.
\newblock volume~24 of \emph{Psychology of Learning and Motivation}, pages 109--165. Academic Press.

\bibitem[{Muennighoff et~al.(2023)Muennighoff, Wang, Sutawika, Roberts, Biderman, Scao, Bari, Shen, Yong, Schoelkopf, Tang, Radev, Aji, Almubarak, Albanie, Alyafeai, Webson, Raff, and Raffel}]{muennighoff2023crosslingual}
Niklas Muennighoff, Thomas Wang, Lintang Sutawika, Adam Roberts, Stella Biderman, Teven~Le Scao, M~Saiful Bari, Sheng Shen, Zheng-Xin Yong, Hailey Schoelkopf, Xiangru Tang, Dragomir Radev, Alham~Fikri Aji, Khalid Almubarak, Samuel Albanie, Zaid Alyafeai, Albert Webson, Edward Raff, and Colin Raffel. 2023.
\newblock \href {https://arxiv.org/abs/2211.01786} {Crosslingual generalization through multitask finetuning}.
\newblock \emph{Preprint}, arXiv:2211.01786.

\bibitem[{Parovi{\'c} et~al.(2022)Parovi{\'c}, Glava{\v{s}}, Vuli{\'c}, and Korhonen}]{parovic-etal-2022-bad}
Marinela Parovi{\'c}, Goran Glava{\v{s}}, Ivan Vuli{\'c}, and Anna Korhonen. 2022.
\newblock \href {https://doi.org/10.18653/v1/2022.naacl-main.130} {{BAD}-{X}: Bilingual adapters improve zero-shot cross-lingual transfer}.
\newblock In \emph{Proceedings of the 2022 Conference of the North American Chapter of the Association for Computational Linguistics: Human Language Technologies}, pages 1791--1799, Seattle, United States. Association for Computational Linguistics.

\bibitem[{Pfeiffer et~al.(2022)Pfeiffer, Goyal, Lin, Li, Cross, Riedel, and Artetxe}]{pfeiffer-etal-2022-lifting}
Jonas Pfeiffer, Naman Goyal, Xi~Lin, Xian Li, James Cross, Sebastian Riedel, and Mikel Artetxe. 2022.
\newblock \href {https://doi.org/10.18653/v1/2022.naacl-main.255} {Lifting the curse of multilinguality by pre-training modular transformers}.
\newblock In \emph{Proceedings of the 2022 Conference of the North American Chapter of the Association for Computational Linguistics: Human Language Technologies}, pages 3479--3495, Seattle, United States. Association for Computational Linguistics.

\bibitem[{Pfeiffer et~al.(2023)Pfeiffer, Piccinno, Nicosia, Wang, Reid, and Ruder}]{pfeiffer-etal-2023-mmt5}
Jonas Pfeiffer, Francesco Piccinno, Massimo Nicosia, Xinyi Wang, Machel Reid, and Sebastian Ruder. 2023.
\newblock \href {https://doi.org/10.18653/v1/2023.findings-emnlp.132} {mm{T}5: Modular multilingual pre-training solves source language hallucinations}.
\newblock In \emph{Findings of the Association for Computational Linguistics: EMNLP 2023}, pages 1978--2008, Singapore. Association for Computational Linguistics.

\bibitem[{Pfeiffer et~al.(2020)Pfeiffer, Vuli{\'c}, Gurevych, and Ruder}]{pfeiffer-etal-2020-mad}
Jonas Pfeiffer, Ivan Vuli{\'c}, Iryna Gurevych, and Sebastian Ruder. 2020.
\newblock \href {https://doi.org/10.18653/v1/2020.emnlp-main.617} {{MAD-X}: {A}n {A}dapter-{B}ased {F}ramework for {M}ulti-{T}ask {C}ross-{L}ingual {T}ransfer}.
\newblock In \emph{Proceedings of the 2020 Conference on Empirical Methods in Natural Language Processing (EMNLP)}, pages 7654--7673, Online. Association for Computational Linguistics.

\bibitem[{Philippy et~al.(2024)Philippy, Guo, Haddadan, Lothritz, Klein, and F.~Bissyand{\'e}}]{philippy-etal-2024-soft}
Fred Philippy, Siwen Guo, Shohreh Haddadan, Cedric Lothritz, Jacques Klein, and Tegawend{\'e} F.~Bissyand{\'e}. 2024.
\newblock \href {https://aclanthology.org/2024.moomin-1.2} {Soft prompt tuning for cross-lingual transfer: When less is more}.
\newblock In \emph{Proceedings of the 1st Workshop on Modular and Open Multilingual NLP (MOOMIN 2024)}, pages 7--15, St Julians, Malta. Association for Computational Linguistics.

\bibitem[{Pikuliak et~al.(2023)Pikuliak, Srba, Moro, Hromadka, Smolen, Melisek, Vykopal, Simko, Podrouzek, and Bielikova}]{pikuliak2023multilingual}
Matúš Pikuliak, Ivan Srba, Robert Moro, Timo Hromadka, Timotej Smolen, Martin Melisek, Ivan Vykopal, Jakub Simko, Juraj Podrouzek, and Maria Bielikova. 2023.
\newblock \href {https://arxiv.org/abs/2305.07991} {Multilingual previously fact-checked claim retrieval}.
\newblock \emph{Preprint}, arXiv:2305.07991.

\bibitem[{Pikuliak et~al.(2021)Pikuliak, Šimko, and Bieliková}]{PIKULIAK2021113765}
Matúš Pikuliak, Marián Šimko, and Mária Bieliková. 2021.
\newblock \href {https://doi.org/10.1016/j.eswa.2020.113765} {Cross-lingual learning for text processing: A survey}.
\newblock \emph{Expert Systems with Applications}, 165:113765.

\bibitem[{Poesina et~al.(2024)Poesina, Caragea, and Ionescu}]{poesina-etal-2024-novel}
Eduard Poesina, Cornelia Caragea, and Radu Ionescu. 2024.
\newblock \href {https://doi.org/10.18653/v1/2024.acl-long.15} {A novel cartography-based curriculum learning method applied on {R}o{NLI}: The first {R}omanian natural language inference corpus}.
\newblock In \emph{Proceedings of the 62nd Annual Meeting of the Association for Computational Linguistics (Volume 1: Long Papers)}, pages 236--253, Bangkok, Thailand. Association for Computational Linguistics.

\bibitem[{Ponti et~al.(2020)Ponti, Glavaš, Majewska, Liu, Vulić, and Korhonen}]{ponti2020xcopa}
Edoardo~Maria Ponti, Goran Glavaš, Olga Majewska, Qianchu Liu, Ivan Vulić, and Anna Korhonen. 2020.
\newblock \href {https://arxiv.org/abs/2005.00333} {Xcopa: A multilingual dataset for causal commonsense reasoning}.
\newblock \emph{Preprint}, arXiv:2005.00333.

\bibitem[{Raffel et~al.(2020)Raffel, Shazeer, Roberts, Lee, Narang, Matena, Zhou, Li, and Liu}]{10.5555/3455716.3455856}
Colin Raffel, Noam Shazeer, Adam Roberts, Katherine Lee, Sharan Narang, Michael Matena, Yanqi Zhou, Wei Li, and Peter~J. Liu. 2020.
\newblock Exploring the limits of transfer learning with a unified text-to-text transformer.
\newblock \emph{J. Mach. Learn. Res.}, 21(1).

\bibitem[{Rahimi et~al.(2019)Rahimi, Li, and Cohn}]{rahimi-etal-2019-massively}
Afshin Rahimi, Yuan Li, and Trevor Cohn. 2019.
\newblock \href {https://doi.org/10.18653/v1/P19-1015} {Massively multilingual transfer for {NER}}.
\newblock In \emph{Proceedings of the 57th Annual Meeting of the Association for Computational Linguistics}, pages 151--164, Florence, Italy. Association for Computational Linguistics.

\bibitem[{Rajpurkar et~al.(2016)Rajpurkar, Zhang, Lopyrev, and Liang}]{rajpurkar-etal-2016-squad}
Pranav Rajpurkar, Jian Zhang, Konstantin Lopyrev, and Percy Liang. 2016.
\newblock \href {https://doi.org/10.18653/v1/D16-1264} {{SQ}u{AD}: 100,000+ questions for machine comprehension of text}.
\newblock In \emph{Proceedings of the 2016 Conference on Empirical Methods in Natural Language Processing}, pages 2383--2392, Austin, Texas. Association for Computational Linguistics.

\bibitem[{Rathore et~al.(2023)Rathore, Dhingra, Singla, and {Mausam}}]{rathore-etal-2023-zgul}
Vipul Rathore, Rajdeep Dhingra, Parag Singla, and {Mausam}. 2023.
\newblock \href {https://doi.org/10.18653/v1/2023.emnlp-main.431} {{ZGUL}: Zero-shot generalization to unseen languages using multi-source ensembling of language adapters}.
\newblock In \emph{Proceedings of the 2023 Conference on Empirical Methods in Natural Language Processing}, pages 6969--6987, Singapore. Association for Computational Linguistics.

\bibitem[{Ren et~al.(2024)Ren, Li, Wang, Zhao, and Qin}]{ren2024analyzing}
Weijieying Ren, Xinlong Li, Lei Wang, Tianxiang Zhao, and Wei Qin. 2024.
\newblock \href {https://arxiv.org/abs/2402.18865} {Analyzing and reducing catastrophic forgetting in parameter efficient tuning}.
\newblock \emph{Preprint}, arXiv:2402.18865.

\bibitem[{Tang et~al.(2020)Tang, Tran, Li, Chen, Goyal, Chaudhary, Gu, and Fan}]{tang2020multilingualtranslationextensiblemultilingual}
Yuqing Tang, Chau Tran, Xian Li, Peng-Jen Chen, Naman Goyal, Vishrav Chaudhary, Jiatao Gu, and Angela Fan. 2020.
\newblock \href {https://arxiv.org/abs/2008.00401} {Multilingual translation with extensible multilingual pretraining and finetuning}.
\newblock \emph{Preprint}, arXiv:2008.00401.

\bibitem[{{\"U}st{\"u}n et~al.(2020){\"U}st{\"u}n, Bisazza, Bouma, and van Noord}]{ustun-etal-2020-udapter}
Ahmet {\"U}st{\"u}n, Arianna Bisazza, Gosse Bouma, and Gertjan van Noord. 2020.
\newblock \href {https://doi.org/10.18653/v1/2020.emnlp-main.180} {{UD}apter: Language adaptation for truly {U}niversal {D}ependency parsing}.
\newblock In \emph{Proceedings of the 2020 Conference on Empirical Methods in Natural Language Processing (EMNLP)}, pages 2302--2315, Online. Association for Computational Linguistics.

\bibitem[{Vemula et~al.(2022)Vemula, Nuthi, and Srivastava}]{vemula-etal-2022-tequad}
Rakesh Vemula, Mani Nuthi, and Manish Srivastava. 2022.
\newblock \href {https://aclanthology.org/2022.icon-main.36} {{T}e{Q}u{AD}:{T}elugu question answering dataset}.
\newblock In \emph{Proceedings of the 19th International Conference on Natural Language Processing (ICON)}, pages 300--307, New Delhi, India. Association for Computational Linguistics.

\bibitem[{Vu et~al.(2022{\natexlab{a}})Vu, Barua, Lester, Cer, Iyyer, and Constant}]{vu-etal-2022-overcoming}
Tu~Vu, Aditya Barua, Brian Lester, Daniel Cer, Mohit Iyyer, and Noah Constant. 2022{\natexlab{a}}.
\newblock \href {https://doi.org/10.18653/v1/2022.emnlp-main.630} {Overcoming catastrophic forgetting in zero-shot cross-lingual generation}.
\newblock In \emph{Proceedings of the 2022 Conference on Empirical Methods in Natural Language Processing}, pages 9279--9300, Abu Dhabi, United Arab Emirates. Association for Computational Linguistics.

\bibitem[{Vu et~al.(2022{\natexlab{b}})Vu, Lester, Constant, Al-Rfou{'}, and Cer}]{vu-etal-2022-spot}
Tu~Vu, Brian Lester, Noah Constant, Rami Al-Rfou{'}, and Daniel Cer. 2022{\natexlab{b}}.
\newblock \href {https://doi.org/10.18653/v1/2022.acl-long.346} {{SP}o{T}: Better frozen model adaptation through soft prompt transfer}.
\newblock In \emph{Proceedings of the 60th Annual Meeting of the Association for Computational Linguistics (Volume 1: Long Papers)}, pages 5039--5059, Dublin, Ireland. Association for Computational Linguistics.

\bibitem[{Wang et~al.(2024)Wang, Wang, Wang, Cao, Saurous, and Kim}]{10.5555/3666122.3668959}
Bailin Wang, Zi~Wang, Xuezhi Wang, Yuan Cao, Rif~A. Saurous, and Yoon Kim. 2024.
\newblock Grammar prompting for domain-specific language generation with large language models.
\newblock In \emph{Proceedings of the 37th International Conference on Neural Information Processing Systems}, NIPS '23, Red Hook, NY, USA. Curran Associates Inc.

\bibitem[{Wanjawa et~al.(2023)Wanjawa, Wanzare, Indede, Mconyango, Muchemi, and Ombui}]{10.1145/3578553}
Barack~W. Wanjawa, Lilian D.~A. Wanzare, Florence Indede, Owen Mconyango, Lawrence Muchemi, and Edward Ombui. 2023.
\newblock \href {https://doi.org/10.1145/3578553} {Kenswquad—a question answering dataset for swahili low-resource language}.
\newblock \emph{ACM Trans. Asian Low-Resour. Lang. Inf. Process.}, 22(4).

\bibitem[{Workshop et~al.(2023)Workshop, :, Scao, Fan, Akiki, Pavlick, Ilić, Hesslow, Castagné, Luccioni, Yvon, Gallé, Tow, Rush, Biderman, Webson, Ammanamanchi, Wang, Sagot, Muennighoff, del Moral, Ruwase, Bawden, Bekman, McMillan-Major, Beltagy, Nguyen, Saulnier, Tan, Suarez, Sanh, Laurençon, Jernite, Launay, Mitchell, Raffel, Gokaslan, Simhi, Soroa, Aji, Alfassy, Rogers, Nitzav, Xu, Mou, Emezue, Klamm, Leong, van Strien, Adelani, Radev, Ponferrada, Levkovizh, Kim, Natan, Toni, Dupont, Kruszewski, Pistilli, Elsahar, Benyamina, Tran, Yu, Abdulmumin, Johnson, Gonzalez-Dios, de~la Rosa, Chim, Dodge, Zhu, Chang, Frohberg, Tobing, Bhattacharjee, Almubarak, Chen, Lo, Werra, Weber, Phan, allal, Tanguy, Dey, Muñoz, Masoud, Grandury, Šaško, Huang, Coavoux, Singh, Jiang, Vu, Jauhar, Ghaleb, Subramani, Kassner, Khamis, Nguyen, Espejel, de~Gibert, Villegas, Henderson, Colombo, Amuok, Lhoest, Harliman, Bommasani, López, Ribeiro, Osei, Pyysalo, Nagel, Bose, Muhammad, Sharma, Longpre, Nikpoor, Silberberg, Pai,
  Zink, Torrent, Schick, Thrush, Danchev, Nikoulina, Laippala, Lepercq, Prabhu, Alyafeai, Talat, Raja, Heinzerling, Si, Taşar, Salesky, Mielke, Lee, Sharma, Santilli, Chaffin, Stiegler, Datta, Szczechla, Chhablani, Wang, Pandey, Strobelt, Fries, Rozen, Gao, Sutawika, Bari, Al-shaibani, Manica, Nayak, Teehan, Albanie, Shen, Ben-David, Bach, Kim, Bers, Fevry, Neeraj, Thakker, Raunak, Tang, Yong, Sun, Brody, Uri, Tojarieh, Roberts, Chung, Tae, Phang, Press, Li, Narayanan, Bourfoune, Casper, Rasley, Ryabinin, Mishra, Zhang, Shoeybi, Peyrounette, Patry, Tazi, Sanseviero, von Platen, Cornette, Lavallée, Lacroix, Rajbhandari, Gandhi, Smith, Requena, Patil, Dettmers, Baruwa, Singh, Cheveleva, Ligozat, Subramonian, Névéol, Lovering, Garrette, Tunuguntla, Reiter, Taktasheva, Voloshina, Bogdanov, Winata, Schoelkopf, Kalo, Novikova, Forde, Clive, Kasai, Kawamura, Hazan, Carpuat, Clinciu, Kim, Cheng, Serikov, Antverg, van~der Wal, Zhang, Zhang, Gehrmann, Mirkin, Pais, Shavrina, Scialom, Yun, Limisiewicz, Rieser,
  Protasov, Mikhailov, Pruksachatkun, Belinkov, Bamberger, Kasner, Rueda, Pestana, Feizpour, Khan, Faranak, Santos, Hevia, Unldreaj, Aghagol, Abdollahi, Tammour, HajiHosseini, Behroozi, Ajibade, Saxena, Ferrandis, McDuff, Contractor, Lansky, David, Kiela, Nguyen, Tan, Baylor, Ozoani, Mirza, Ononiwu, Rezanejad, Jones, Bhattacharya, Solaiman, Sedenko, Nejadgholi, Passmore, Seltzer, Sanz, Dutra, Samagaio, Elbadri, Mieskes, Gerchick, Akinlolu, McKenna, Qiu, Ghauri, Burynok, Abrar, Rajani, Elkott, Fahmy, Samuel, An, Kromann, Hao, Alizadeh, Shubber, Wang, Roy, Viguier, Le, Oyebade, Le, Yang, Nguyen, Kashyap, Palasciano, Callahan, Shukla, Miranda-Escalada, Singh, Beilharz, Wang, Brito, Zhou, Jain, Xu, Fourrier, Periñán, Molano, Yu, Manjavacas, Barth, Fuhrimann, Altay, Bayrak, Burns, Vrabec, Bello, Dash, Kang, Giorgi, Golde, Posada, Sivaraman, Bulchandani, Liu, Shinzato, de~Bykhovetz, Takeuchi, Pàmies, Castillo, Nezhurina, Sänger, Samwald, Cullan, Weinberg, Wolf, Mihaljcic, Liu, Freidank, Kang, Seelam, Dahlberg,
  Broad, Muellner, Fung, Haller, Chandrasekhar, Eisenberg, Martin, Canalli, Su, Su, Cahyawijaya, Garda, Deshmukh, Mishra, Kiblawi, Ott, Sang-aroonsiri, Kumar, Schweter, Bharati, Laud, Gigant, Kainuma, Kusa, Labrak, Bajaj, Venkatraman, Xu, Xu, Xu, Tan, Xie, Ye, Bras, Belkada, and Wolf}]{workshop2023bloom}
BigScience Workshop, :, Teven~Le Scao, Angela Fan, Christopher Akiki, Ellie Pavlick, Suzana Ilić, Daniel Hesslow, Roman Castagné, Alexandra~Sasha Luccioni, François Yvon, Matthias Gallé, Jonathan Tow, Alexander~M. Rush, Stella Biderman, Albert Webson, Pawan~Sasanka Ammanamanchi, Thomas Wang, Benoît Sagot, Niklas Muennighoff, Albert~Villanova del Moral, Olatunji Ruwase, Rachel Bawden, Stas Bekman, Angelina McMillan-Major, Iz~Beltagy, Huu Nguyen, Lucile Saulnier, Samson Tan, Pedro~Ortiz Suarez, Victor Sanh, Hugo Laurençon, Yacine Jernite, Julien Launay, Margaret Mitchell, Colin Raffel, Aaron Gokaslan, Adi Simhi, Aitor Soroa, Alham~Fikri Aji, Amit Alfassy, Anna Rogers, Ariel~Kreisberg Nitzav, Canwen Xu, Chenghao Mou, Chris Emezue, Christopher Klamm, Colin Leong, Daniel van Strien, David~Ifeoluwa Adelani, Dragomir Radev, Eduardo~González Ponferrada, Efrat Levkovizh, Ethan Kim, Eyal~Bar Natan, Francesco~De Toni, Gérard Dupont, Germán Kruszewski, Giada Pistilli, Hady Elsahar, Hamza Benyamina, Hieu Tran,
  Ian Yu, Idris Abdulmumin, Isaac Johnson, Itziar Gonzalez-Dios, Javier de~la Rosa, Jenny Chim, Jesse Dodge, Jian Zhu, Jonathan Chang, Jörg Frohberg, Joseph Tobing, Joydeep Bhattacharjee, Khalid Almubarak, Kimbo Chen, Kyle Lo, Leandro~Von Werra, Leon Weber, Long Phan, Loubna~Ben allal, Ludovic Tanguy, Manan Dey, Manuel~Romero Muñoz, Maraim Masoud, María Grandury, Mario Šaško, Max Huang, Maximin Coavoux, Mayank Singh, Mike Tian-Jian Jiang, Minh~Chien Vu, Mohammad~A. Jauhar, Mustafa Ghaleb, Nishant Subramani, Nora Kassner, Nurulaqilla Khamis, Olivier Nguyen, Omar Espejel, Ona de~Gibert, Paulo Villegas, Peter Henderson, Pierre Colombo, Priscilla Amuok, Quentin Lhoest, Rheza Harliman, Rishi Bommasani, Roberto~Luis López, Rui Ribeiro, Salomey Osei, Sampo Pyysalo, Sebastian Nagel, Shamik Bose, Shamsuddeen~Hassan Muhammad, Shanya Sharma, Shayne Longpre, Somaieh Nikpoor, Stanislav Silberberg, Suhas Pai, Sydney Zink, Tiago~Timponi Torrent, Timo Schick, Tristan Thrush, Valentin Danchev, Vassilina Nikoulina,
  Veronika Laippala, Violette Lepercq, Vrinda Prabhu, Zaid Alyafeai, Zeerak Talat, Arun Raja, Benjamin Heinzerling, Chenglei Si, Davut~Emre Taşar, Elizabeth Salesky, Sabrina~J. Mielke, Wilson~Y. Lee, Abheesht Sharma, Andrea Santilli, Antoine Chaffin, Arnaud Stiegler, Debajyoti Datta, Eliza Szczechla, Gunjan Chhablani, Han Wang, Harshit Pandey, Hendrik Strobelt, Jason~Alan Fries, Jos Rozen, Leo Gao, Lintang Sutawika, M~Saiful Bari, Maged~S. Al-shaibani, Matteo Manica, Nihal Nayak, Ryan Teehan, Samuel Albanie, Sheng Shen, Srulik Ben-David, Stephen~H. Bach, Taewoon Kim, Tali Bers, Thibault Fevry, Trishala Neeraj, Urmish Thakker, Vikas Raunak, Xiangru Tang, Zheng-Xin Yong, Zhiqing Sun, Shaked Brody, Yallow Uri, Hadar Tojarieh, Adam Roberts, Hyung~Won Chung, Jaesung Tae, Jason Phang, Ofir Press, Conglong Li, Deepak Narayanan, Hatim Bourfoune, Jared Casper, Jeff Rasley, Max Ryabinin, Mayank Mishra, Minjia Zhang, Mohammad Shoeybi, Myriam Peyrounette, Nicolas Patry, Nouamane Tazi, Omar Sanseviero, Patrick von
  Platen, Pierre Cornette, Pierre~François Lavallée, Rémi Lacroix, Samyam Rajbhandari, Sanchit Gandhi, Shaden Smith, Stéphane Requena, Suraj Patil, Tim Dettmers, Ahmed Baruwa, Amanpreet Singh, Anastasia Cheveleva, Anne-Laure Ligozat, Arjun Subramonian, Aurélie Névéol, Charles Lovering, Dan Garrette, Deepak Tunuguntla, Ehud Reiter, Ekaterina Taktasheva, Ekaterina Voloshina, Eli Bogdanov, Genta~Indra Winata, Hailey Schoelkopf, Jan-Christoph Kalo, Jekaterina Novikova, Jessica~Zosa Forde, Jordan Clive, Jungo Kasai, Ken Kawamura, Liam Hazan, Marine Carpuat, Miruna Clinciu, Najoung Kim, Newton Cheng, Oleg Serikov, Omer Antverg, Oskar van~der Wal, Rui Zhang, Ruochen Zhang, Sebastian Gehrmann, Shachar Mirkin, Shani Pais, Tatiana Shavrina, Thomas Scialom, Tian Yun, Tomasz Limisiewicz, Verena Rieser, Vitaly Protasov, Vladislav Mikhailov, Yada Pruksachatkun, Yonatan Belinkov, Zachary Bamberger, Zdeněk Kasner, Alice Rueda, Amanda Pestana, Amir Feizpour, Ammar Khan, Amy Faranak, Ana Santos, Anthony Hevia, Antigona
  Unldreaj, Arash Aghagol, Arezoo Abdollahi, Aycha Tammour, Azadeh HajiHosseini, Bahareh Behroozi, Benjamin Ajibade, Bharat Saxena, Carlos~Muñoz Ferrandis, Daniel McDuff, Danish Contractor, David Lansky, Davis David, Douwe Kiela, Duong~A. Nguyen, Edward Tan, Emi Baylor, Ezinwanne Ozoani, Fatima Mirza, Frankline Ononiwu, Habib Rezanejad, Hessie Jones, Indrani Bhattacharya, Irene Solaiman, Irina Sedenko, Isar Nejadgholi, Jesse Passmore, Josh Seltzer, Julio~Bonis Sanz, Livia Dutra, Mairon Samagaio, Maraim Elbadri, Margot Mieskes, Marissa Gerchick, Martha Akinlolu, Michael McKenna, Mike Qiu, Muhammed Ghauri, Mykola Burynok, Nafis Abrar, Nazneen Rajani, Nour Elkott, Nour Fahmy, Olanrewaju Samuel, Ran An, Rasmus Kromann, Ryan Hao, Samira Alizadeh, Sarmad Shubber, Silas Wang, Sourav Roy, Sylvain Viguier, Thanh Le, Tobi Oyebade, Trieu Le, Yoyo Yang, Zach Nguyen, Abhinav~Ramesh Kashyap, Alfredo Palasciano, Alison Callahan, Anima Shukla, Antonio Miranda-Escalada, Ayush Singh, Benjamin Beilharz, Bo~Wang, Caio Brito,
  Chenxi Zhou, Chirag Jain, Chuxin Xu, Clémentine Fourrier, Daniel~León Periñán, Daniel Molano, Dian Yu, Enrique Manjavacas, Fabio Barth, Florian Fuhrimann, Gabriel Altay, Giyaseddin Bayrak, Gully Burns, Helena~U. Vrabec, Imane Bello, Ishani Dash, Jihyun Kang, John Giorgi, Jonas Golde, Jose~David Posada, Karthik~Rangasai Sivaraman, Lokesh Bulchandani, Lu~Liu, Luisa Shinzato, Madeleine~Hahn de~Bykhovetz, Maiko Takeuchi, Marc Pàmies, Maria~A Castillo, Marianna Nezhurina, Mario Sänger, Matthias Samwald, Michael Cullan, Michael Weinberg, Michiel~De Wolf, Mina Mihaljcic, Minna Liu, Moritz Freidank, Myungsun Kang, Natasha Seelam, Nathan Dahlberg, Nicholas~Michio Broad, Nikolaus Muellner, Pascale Fung, Patrick Haller, Ramya Chandrasekhar, Renata Eisenberg, Robert Martin, Rodrigo Canalli, Rosaline Su, Ruisi Su, Samuel Cahyawijaya, Samuele Garda, Shlok~S Deshmukh, Shubhanshu Mishra, Sid Kiblawi, Simon Ott, Sinee Sang-aroonsiri, Srishti Kumar, Stefan Schweter, Sushil Bharati, Tanmay Laud, Théo Gigant, Tomoya
  Kainuma, Wojciech Kusa, Yanis Labrak, Yash~Shailesh Bajaj, Yash Venkatraman, Yifan Xu, Yingxin Xu, Yu~Xu, Zhe Tan, Zhongli Xie, Zifan Ye, Mathilde Bras, Younes Belkada, and Thomas Wolf. 2023.
\newblock \href {https://arxiv.org/abs/2211.05100} {Bloom: A 176b-parameter open-access multilingual language model}.
\newblock \emph{Preprint}, arXiv:2211.05100.

\bibitem[{Wu and Dredze(2020)}]{wu2020languages}
Shijie Wu and Mark Dredze. 2020.
\newblock \href {https://arxiv.org/abs/2005.09093} {Are all languages created equal in multilingual bert?}
\newblock \emph{Preprint}, arXiv:2005.09093.

\bibitem[{Xie et~al.(2024)Xie, Zhao, Yu, and Li}]{xie2024discovering}
Zhihui Xie, Handong Zhao, Tong Yu, and Shuai Li. 2024.
\newblock \href {https://arxiv.org/abs/2401.05792} {Discovering low-rank subspaces for language-agnostic multilingual representations}.
\newblock \emph{Preprint}, arXiv:2401.05792.

\bibitem[{Xu et~al.(2023)Xu, Xie, Qin, Tao, and Wang}]{xu2023parameterefficient}
Lingling Xu, Haoran Xie, Si-Zhao~Joe Qin, Xiaohui Tao, and Fu~Lee Wang. 2023.
\newblock \href {https://arxiv.org/abs/2312.12148} {Parameter-efficient fine-tuning methods for pretrained language models: A critical review and assessment}.
\newblock \emph{Preprint}, arXiv:2312.12148.

\bibitem[{Xue et~al.(2021)Xue, Constant, Roberts, Kale, Al-Rfou, Siddhant, Barua, and Raffel}]{xue2021mt5}
Linting Xue, Noah Constant, Adam Roberts, Mihir Kale, Rami Al-Rfou, Aditya Siddhant, Aditya Barua, and Colin Raffel. 2021.
\newblock \href {https://arxiv.org/abs/2010.11934} {mt5: A massively multilingual pre-trained text-to-text transformer}.
\newblock \emph{Preprint}, arXiv:2010.11934.

\bibitem[{Zhang et~al.(2023)Zhang, Chen, Bukharin, Karampatziakis, He, Cheng, Chen, and Zhao}]{zhang2023adalora}
Qingru Zhang, Minshuo Chen, Alexander Bukharin, Nikos Karampatziakis, Pengcheng He, Yu~Cheng, Weizhu Chen, and Tuo Zhao. 2023.
\newblock \href {https://arxiv.org/abs/2303.10512} {Adalora: Adaptive budget allocation for parameter-efficient fine-tuning}.
\newblock \emph{Preprint}, arXiv:2303.10512.

\bibitem[{Üstün et~al.(2024)Üstün, Aryabumi, Yong, Ko, D'souza, Onilude, Bhandari, Singh, Ooi, Kayid, Vargus, Blunsom, Longpre, Muennighoff, Fadaee, Kreutzer, and Hooker}]{ustun2024aya}
Ahmet Üstün, Viraat Aryabumi, Zheng-Xin Yong, Wei-Yin Ko, Daniel D'souza, Gbemileke Onilude, Neel Bhandari, Shivalika Singh, Hui-Lee Ooi, Amr Kayid, Freddie Vargus, Phil Blunsom, Shayne Longpre, Niklas Muennighoff, Marzieh Fadaee, Julia Kreutzer, and Sara Hooker. 2024.
\newblock \href {https://arxiv.org/abs/2402.07827} {Aya model: An instruction finetuned open-access multilingual language model}.
\newblock \emph{Preprint}, arXiv:2402.07827.

\end{thebibliography}
